\RequirePackage{snapshot}
\documentclass[letterpaper, 10 pt, conference]{style/ieeeconf}
\IEEEoverridecommandlockouts
\usepackage{amsmath,amssymb}
\usepackage{bm}
\usepackage{palatino}
\usepackage[normalem]{ulem}

\usepackage[english]{babel}
\usepackage{xcolor}
\usepackage{algorithm}
\usepackage[]{algpseudocode}
\usepackage{array}
\usepackage{url}
\usepackage{color}
\usepackage[tight,footnotesize]{subfigure}
\usepackage[utf8]{inputenc}
\usepackage{cite}
\usepackage{multicol}
\usepackage{graphics,graphicx} % for pdf, bitmapped graphics files
\usepackage{epsfig} % for postscript graphics files
\usepackage{mathptmx} % assume new font selection scheme installed
\usepackage{array}

\newcommand{\Rnum}{\mathbb{R}} % Symbol fo the real numbers set

\newcommand{\degree}{\ensuremath{^\circ}}				% define the degree symbol
\newcommand{\mat}[1]{\ensuremath{\begin{bmatrix}#1\end{bmatrix}}}	% matrix
\DeclareMathAlphabet{\mathcal}{OMS}{zplm}{m}{n}

\definecolor{darkGreen}{RGB}{0,102,0}
\definecolor{lightGrey}{RGB}{96,96,96}

\usepackage{glossaries}
	\newacronym{hyq}{HyQ}{Hydraulically actuated Quadruped}
	\newacronym{KBF}{KBF}{Knee Bent Forward}
	\newacronym{KBB}{KBB}{Knee Bent Backward}
	\newacronym{mbdo}{MBDO}{Momentum Based Disturbance Observer}
	\newacronym{prmpc}{PR-MPC}{Policy-Regularized Model Predictive Control}
	\newacronym{mpc}{MPC}{Model Predictive Control}
	\newacronym{dof}{DoF}{Degree of Freedom}
	\newacronym{eom}{EoM}{Equation of Motion}
    \newacronym{IMU}{imu}{Inertial Measurement Unit}
	\newacronym{com}{CoM}{Center of Mass}
	\newacronym{cop}{CoP}{Center of Pressure}
	\newacronym{grf}{GRF}{Ground Reaction Force}
	\newacronym{zmp}{ZMP}{Zero Moment Point}
	\newacronym{lip}{LIP}{Linear Inverted Pendulum}
	\newacronym{HAA}{HAA}{Hip Abduction/Adduction}
	\newacronym{ccw}{CCW}{Counter Clock Wise}
	\newacronym{slerp}{SLERP}{Spherical Linear intERPolation}
\usepackage[tight]{units}

\newcommand{\eref}[1]{Eq.~(\ref{#1})}

\DeclareMathOperator{\diag}{diag}

\newcommand{\eg}{\emph{e.g.~}}
\newcommand{\etal}{\emph{et al.~}}
\newcommand{\ie}{\emph{i.e.~}}
\usepackage[version-1-compatibility]{siunitx}
\DeclareMathOperator{\atantwo}{atan2}
\DeclareMathOperator{\atan}{atan}
\usepackage[colorlinks,bookmarksopen,bookmarksnumbered,citecolor=red,
urlcolor=red]{hyperref}
\hypersetup
{
	pdftitle = {Heuristic Planning for Rough Terrain Locomotion in Presence
of External Disturbances and Variable Perception Quality},
	pdfauthor = {Michele Focchi},
	pdfsubject = {},
	pdfkeywords = {Legged Robots},
	pdftoolbar = true,
	colorlinks = true,
	linkcolor = blue, %TODO black
	citecolor = green,
	urlcolor = black,
}

\begin{document}
\title{\LARGE \bf Heuristic Planning for Rough Terrain Locomotion in Presence of
External Disturbances and Variable Perception Quality}

\author{{\centering
Michele Focchi$^1$ \quad Romeo Orsolino$^1$ \quad Marco Camurri$^{1,2}$ }\\
\quad Victor Barasuol$^1$ \quad Carlos Mastalli$^{1,3}$ \quad Darwin G.
Caldwell$^1$ \quad Claudio Semini$^1$
\thanks{
\hspace{-1em}$^1$Dynamic Legged Systems lab, Istituto Italiano di
Tecnologia, Genova, Italy. \textit{Email}:
\{\href{mailto:michele.focchi@iit.it}{michele.focchi},
\href{mailto:romeo.orsolino@iit.it}{romeo.orsolino},
\href{mailto:victor.barasuol@iit.it}{victor.barasuol},
\href{mailto:darwin.caldwell@iit.it}{darwin.caldwell},
\href{mailto:claudio.semini@iit.it}{claudio.semini}\}@iit.it.
\newline
$^2$Oxford Robotics Institute, University of Oxford, Oxford, UK.
\newline \textit{Email}:
\href{mailto:mcamurri@robots.ox.ac.uk}{mcamurri@robots.ox.ac.uk}.
\newline
$^3$LAAS-CNRS, Toulouse, France. \textit{Email}:
\href{mailto:carlos.mastalli@laas.fr}{carlos.mastalli@laas.fr}}}

% make the title area
\maketitle
\thispagestyle{empty}
\pagestyle{empty}

\setlength{\medmuskip}{2\medmuskip}% Formerly 4.0mu plus 2.0mu minus 4.0mu -> 4.0mu
\setlength{\thickmuskip}{2\thickmuskip}% Formerly 5.0mu plus 5.0mu -> 5.0mu
\setlength{\thinmuskip}{1\thinmuskip}% Formerly 5.0mu plus 5.0mu -> 5.0mu

\begin{abstract}
The quality of visual feedback can vary
significantly on a legged robot meant to traverse unknown and unstructured
terrains.
The map of the environment, acquired with online state-of-the-art
algorithms, often degrades after a few steps,
due to sensing inaccuracies, slippage and  unexpected disturbances.
If a locomotion algorithm is not designed to deal with
this degradation,
its planned trajectories might end-up to be inconsistent in reality.
In this work, we propose a heuristic-based planning
approach that enables a quadruped robot to successfully
traverse a \textit{significantly}
rough terrain (\eg stones up to \SI{10}{\centi\meter} of diameter),
in \textit{absence} of visual feedback.
When available, the approach allows also to leverage
the visual feedback (\eg to enhance the stepping strategy) in \textit{multiple
ways}, according to the \textit{quality} of the 3D map.
The proposed framework also includes reflexes, triggered in specific situations,
and the possibility to  estimate \textit{online} an unknown time-varying
disturbance and compensate for it.
We demonstrate the effectiveness of the  approach with
experiments performed on our quadruped  robot \emph{HyQ} (\SI{85}{\kilo\gram}),
traversing
 different terrains, such as: ramps, rocks, bricks, pallets and  stairs.
We also demonstrate the  capability to estimate and compensate for external disturbances
by showing the robot walking up a ramp while pulling a cart attached to its
back.
\end{abstract}

%\tableofcontents

\section{Introduction}
Legged robots are mainly designed to traverse unstructured environments, which
are often demanding in terms of torques and speeds.
%TO motivation
Trajectory optimization
\cite{Bretl2008,Pardo2015a,mastalli17icra,Winkler2018,Aceituno-Cabezas2018} can
be used to generate dynamically stable body motions, taking into account:
robot dynamics, kinematic limits, leg reachability for motion generation and
foothold selection.
In particular, motion planning enables the necessary \textit{anticipative}
behaviors to address appropriately the terrain. These include: obstacle
avoidance, foothold selection, contact force generation for optimal body motion
and goal-driven navigation (\eg \cite{mastalli15tepra,mastalli17icra}).
%limitations of the planning solution
Furthermore, when the flat terrain assumption is no longer valid,
a $3D/2.5D$  map of the environment is required, to appropriately
address  uneven  terrain morphology, through  optimization.

%BD old shitty story
%In the last few years, robots from Boston Dynamics demonstrated notable
%examples of dynamic locomotion with a variety of gaits, as well as manipulation
%capabilities;
%however details on these methods are not yet available to the public.

% %full blown TO is not possible in real time
Despite the considerable efforts and recent advances
in this field \cite{Dai2016b, Winkler2018, mastalli17icra, Ponton2017},
optimal planning that takes into account terrain conditions
is still far from being executed \textit{online} on a real platform,
due to the computational complexity involved in the optimization.
These optimization problems are strongly nonlinear, prone to local minima,
and require a significant amount of computation time to be solved
\cite{Dai2016b}.
Some improvements have been recently achieved by computing  convex
approximations
of the terrain \cite{Aceituno-Cabezas2018} or
 the dynamics \cite{Ponton2017}.

Most of these approaches optimize for a whole time
horizon and then execute the motion in an ``open loop'' fashion, rather than optimizing
\textit{online}. Depending on the complexity of terrains and gaits,
Aceituno \etal \cite{Aceituno-Cabezas2018} managed to reduce the computation time
(for a locomotion cycle) in a range between 0.5 $s$ and 1.5  $s$, while the
approach by Ponton
\etal \cite{Ponton2017} requires \SIrange{0.8}{5}{\second}
to optimize for a \SI{10}{\second} horizon. Therefore, it is still not
possible
to optimize \textit{online} and perform replanning (\eg through a \gls*{mpc}
strategy).
%
%accumulated errors
%Moreover, current approaches are limited by the inability to cope

We believe that replanning is a crucial feature to intrinsically cope
with the problem of error accumulation in real scenarios.
These errors can be caused by delays, inaccuracy of the 3D map,
unforeseen events (external pushes, slippages, rock falling), or dynamically
changing and deformable terrains (\eg rolling stones, mud, etc.).
%categorize errors
%In particular, we can categorize the errors as \textit{extrinsic}
%or \textit{intrinsic}. Intrinsic errors include: 

The sources of errors in locomotion are many: a premature/delayed touchdown
due to a change in the terrain inclination, external disturbances, wrong terrain
detection. Other source of errors include: tracking delays in the controller, sensor
calibration errors, filtering delays, offsets, structure compliance, unmodeled
friction and modeling errors in general.
These errors can make the actual robot state diverge from the original plan.
Moreover, in the prospect of tele-operated robots, the user might want to modify
the robot velocity during  locomotion, and this should be reflected in a
prompt change in the motion plan.

% Momentum Based Disturbance Observer
As mentioned, a significant source of errors can come from non-modeled
disturbances, such as external pushes.
% in what we improve wrt rotella (online compensation)
%thesis rotella 4.4.4 (pag 63)  open loop estimation 4.5.2 closed loop exps (pag
%65) %to improve show the estimation during the walk
A possible solution to this particular issue is implementing a disturbance
observer \cite{Stephens11}.
Englsberger \etal \cite{englsberger17iros} proposed a \gls*{mbdo} for
pure linear force disturbances. In his thesis, Rotella and \etal
\cite{rotella18thesis}, implemented an external wrench estimator for a humanoid
robot, based on an Extended Kalman Filter. Besides that, he was also compensating for
the estimated wrench in an inverse dynamics controller. However the experimental
tests on the real robot were limited to a static load, acting on the legs while
performing a switch of contact between the two feet. In a separated
experiment, he also tested against impulsive disturbance without any contact
change. We extend
this work by presenting a \gls*{mbdo} capable of estimating both linear and
angular components of external disturbance  wrenches, of variable
amplitude,
applied on the robot during the execution of a walk.

Even though a disturbance observer can improve the
tracking against external disturbances,
 \textit{online replanning} is still required for rough terrain
locomotion, because it constitutes the basic mechanism to adapt
to the terrain (\textit{terrain adaptation}),  promptly
recover from planning errors and handle 
environmental changes, while simultaneously accommodate for the user set-point.
In particular, the planning horizon
should be large enough to execute the
necessary \textit{anticipative} body motions,
but at the same time the replanning frequency should be high
enough to mitigate the accumulation of errors.
%

%how replanning has been implemented before
Bledt \etal  \cite{wensigHeuristics2017} implemented online replanning,
by introducing a policy-regularized model-predictive control (PR-MPC)
for gait generation, where  a heuristic policy provides additional information
for  the solution of the \gls*{mpc}
problem. The authors found that regularizing the optimization with a policy  it 
improves the cost landscapes and decreases the computation time.
However, this work has not been demonstrated experimentally.
%
%
%MPC

A \gls*{mpc}-based approach has been successfully implemented for a real
humanoid in the past. In his seminal work, Wieber \cite{Wieber2006}
addressed the problem of strong perturbations and tracking errors, yet limiting
the
application to flat terrains.

In the quadruped domain, Cheetah 2 has shown
running/jumping motions over challenging terrains using a \gls*{mpc} controller
\cite{Park2015a}.
More recently, Bellicoso \etal \cite{bellicoso2016, Bellicoso2017}
implemented a 1-step replanning strategy, based on \gls*{zmp} optimization,
on the quadruped robot ANYmal.

\begin{figure}[tb]
\centering
\includegraphics[width=0.8\columnwidth]{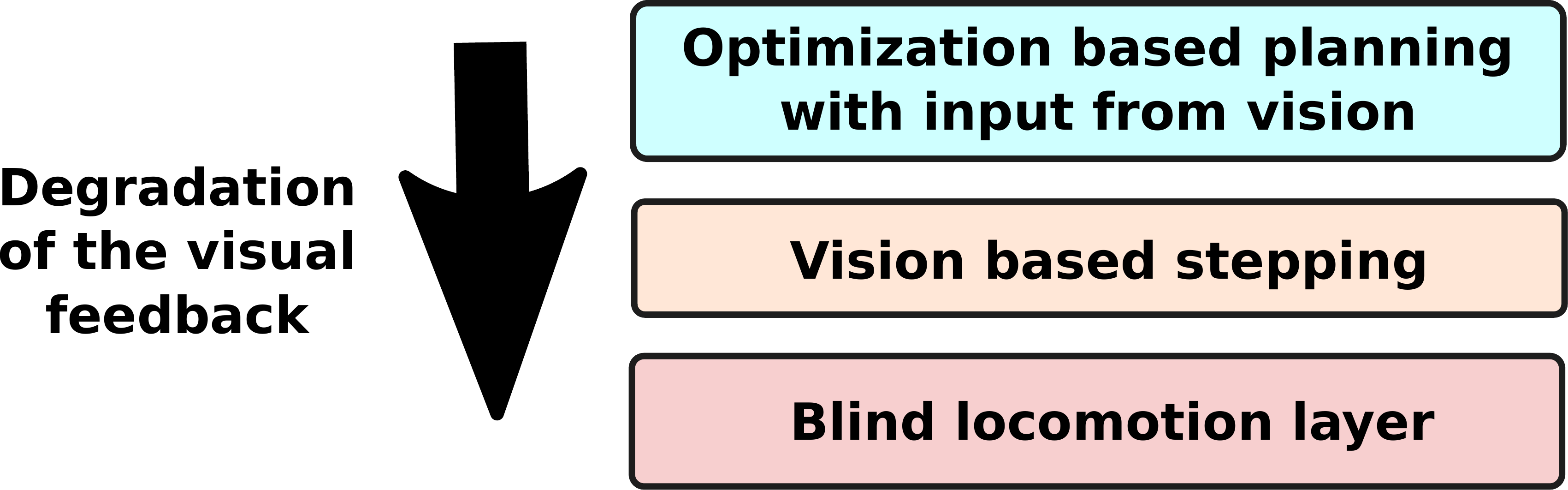}
\caption{Locomotion layers for different quality of the visual feedback.
The bottom layer is purely reactive (blind) locomotion. The vision-based
stepping allows the robot to quickly adapt to local terrain conditions. Then,
the motion planning provides the level of anticipation needed in locomotion over
challenging terrain.}
\label{fig:locomotionLayers}
\end{figure}

%people rely on accurate vision
Most vision-based approaches \cite{mastalli15tepra,mastalli17icra,fankhauser18icra} require reasonably
accurate $3D$ maps of the environment \cite{fallon15}. However, the accuracy of
a $3D$ map strongly depends on reliable state estimation \cite{camurri17ral,
Bloesch2012rss, xinjilefu14icra}, which involves complex sensor fusion
algorithms (inertial, odometry, LiDAR, cameras). Typically, these algorithms
involve the fusion of high-frequency proprioceptive estimates from inertial and
leg odometry \cite{camurri17ral}, with low-frequency pose correction from visual
odometry \cite{Nobili2017}. The proprioceptive estimate (and, indirectly, the
pose correction) can be jeopardized by unforeseen events such as: slippage,
unstable footholds, terrain deformation, and compliance of the mechanical
structure. For the above reasons, we envision different locomotion layers,
according to the quality of the perception feedback available, as depicted in
Fig. \ref{fig:locomotionLayers}.

At the bottom layer we have a \textit{blind} reactive strategy, always
active. This strategy does not require any vision feedback. The layer mainly
contains basic terrain adaptation mechanisms and reflex strategies. Motion
primitives are triggered to override the planner in situations
where the robot could be damaged (\eg stumbling).

In cases when the vision feedback is denied (\eg foggy, smoky, poorly
lit
areas), a \textit{reactive strategy} is preferred \cite{Barasuol2013},
\cite{Focchi2018}. On the other hand, when a degraded visual feedback
is available, this can still be usefully exploited for 1-step horizon planning
(middle block in Fig. \ref{fig:locomotionLayers}) \cite{barasuol15iros}.
When the vision feedback is instead reliable, it can be used for more
sophisticated
terrain assessment \cite{belter11jfr,mastalli18}.
%making it not suitable to be used as an input for a long horizon optimization.

%our contribution
In this work, we present a motion control framework for rough terrain locomotion.
The terrain adaptation and the mitigation of tracking errors is achieved through
a 1-step \textit{online}  replanning strategy, which
can work either blindly or with visual feedback.

%connection to auro
This work builds upon a previously
presented statically stable \textit{crawl} gait \cite{focchi2017auro},
where the main focus was on whole-body control.
Hereby, our focus is mainly on the capability to
adapt to rough terrains during locomotion.
%reflexes incorporated
We incorporate a number of features to increase the
robustness of the locomotion, such as slip detection and two reflex strategies,
previously presented in \cite{focchi17iros}, \cite{focchi2013clawar}. The
reflexes are automatically enabled to address specific situations such as: loss
of mobility (in cases of abrupt terrain changes, see \textit{height reflex} in
\cite{focchi17iros}) and unforeseen frontal impacts (see \textit{step reflex}
in \cite{focchi2013clawar}).

\subsection{Contribution}
\begin{figure}[tb]
\centering
\includegraphics[width=0.59\columnwidth]{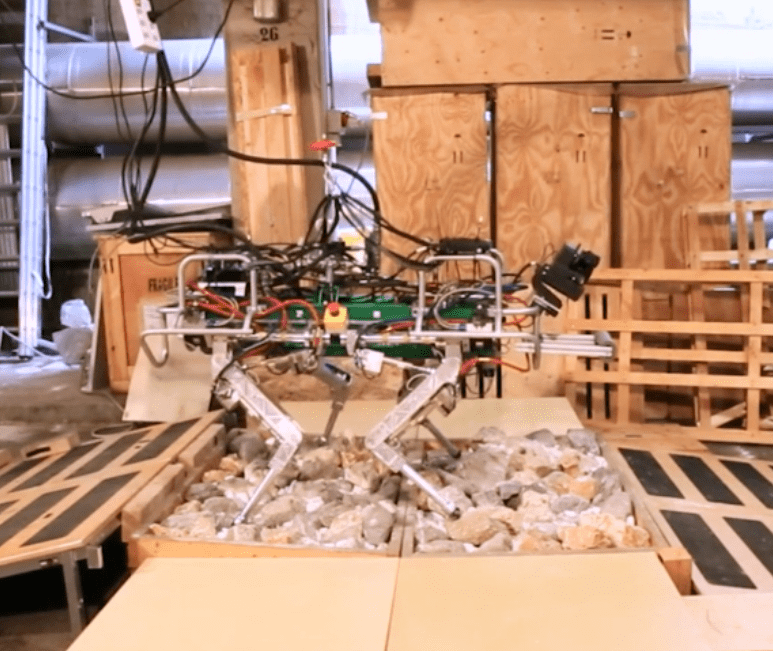}
\includegraphics[width=0.39\columnwidth]{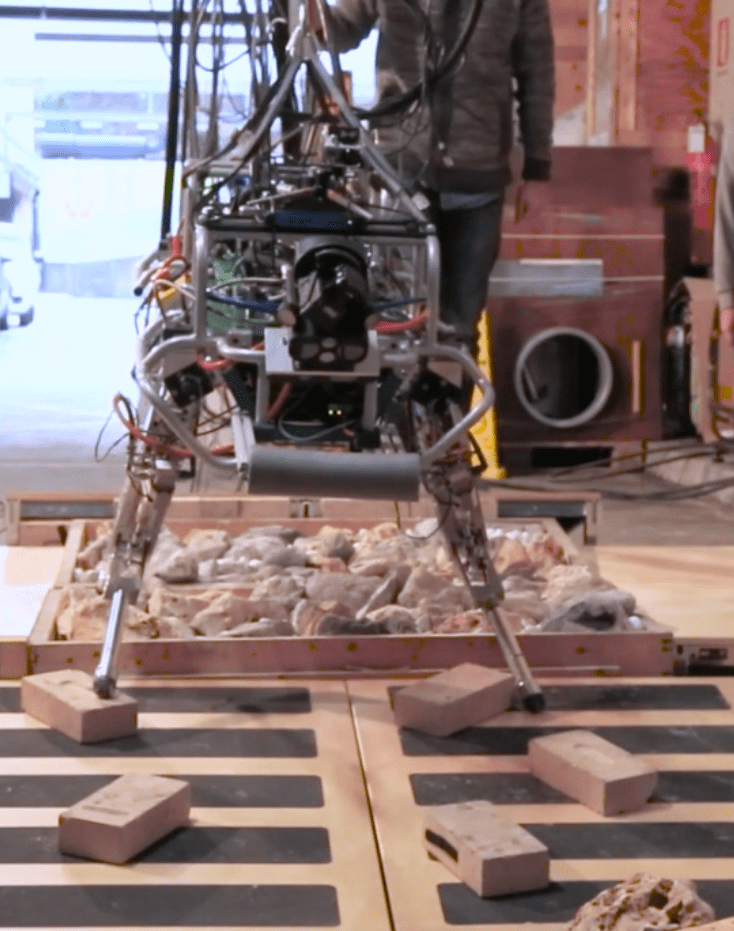}
\caption{\gls{hyq} crawling on a rough terrain playground: (left) lateral and
(right) frontal view. Our locomotion framework can deal with moving rocks and
various terrain elevation changes.}
\vspace{-0.5cm}
\label{fig:roughTerrain}
\end{figure}

%1 contribution - experiments
The main contributions of this article are \textit{experimental}:
\begin{itemize}
	\item We show that with the proposed approach the \gls*{hyq} robot
	\cite{semini2011} can successfully negotiate different types of terrain
	templates (ramps, debris, stairs, steps), some of them with a significant
	roughness\footnote{See accompaning video of rough terrain experiments:
	\url{https://youtu.be/_7ud4zIt-Gw}}
	(Fig.~\ref{fig:roughTerrain}). The size of the stones are up to
	\SI{12}{\centi\meter} (diameter) that is about $26 \%$ of the retractable leg
	length. We also applied this  strategy, with little variations (see Section
	\ref{sec:stairLocomotionMode}), to the task of climbing up and down industrial-size stairs (14cm x 48cm), also performing \SI{90}{\degree} turns while
	climbing the stairs 	in simulation.
	%previous stair climbing
	%In a work \cite{Fankhauser2016} showed the quadruped robot
	%ANYmal climbing industrial stairs (inclination \SI{> 50}{\degree})
	%with the help of its own belly.
	%Differently from their strategy, we propose an omni-directional stair
	%climbing approach which is generic and extends to different step elevations
	%cases.
	%2 contribution
	\item We validate \textit{experimentally} our \gls*{mbdo} for
quadrupedal locomotion. We are able to
	compensate for the disturbances \textit{online} and in close-loop
during locomotion.
	Additionally, while planning the trajectories, we also consider the shift of
	the \gls*{zmp} due to the estimated external disturbance. We show
	\gls*{hyq} walking up a \SI{22}{\degree} inclined ramp, while
pulling a \SI{12}{\kilo\gram}
	wheelbarrow attached to his back with a rope. The wheelbarrow is also
	impulsively loaded up to \SI{15}{\kilo\gram} of extra payload,
incrementally added during the experiment. The robot is able to
	crawl robustly on a flat terrain while leaning against a constant horizontal
	pulling force of \SI{75}{\newton}. To the authors knowledge, this is
the first
	time such tasks are executed on a real quadruped platform.
	%We also demonstrated the effectiveness of our approach in experiments
	%showing HyQ both pulling a cart while walking up  a slope
	%and pulling an horizontal load.

	\item As a marginal contribution, we introduce a \textit{smart terrain
	estimation} algorithm, which improves the state-of-the-art terrain
estimation
	and adaptation of \cite{bellicoso2016}. This is particularly beneficial
in some
	specific situations (\eg when one stance foot is considerably far from
the plane
	fitted by the other three stance feet).
\end{itemize}
%
%This  with very
%large height variations and removes undesired behaviors that might arise
%from other terrain estimation strategies such as the \textit{vertical fit}
%and the \textit{affine fit} methods.\\}
%

\subsection{Echord++}
This work is part of the Echord++ HyQ-REAL experiment, where
novel quadrupedal locomotion strategies are being developed to be used
on newly designed hydraulic robots. At the time of writing this manuscript,
the HyQ-REAL robot was not yet fully assembled. Therefore, the framework is
demonstrated on its predecessor version, HyQ. The major
improvements of HyQ-REAL over HyQ are: more powerful actuators, full power
autonomy (no
tethers), greater range of motion, self-righting capabilities, fully enclosed
chassis. The
presented software framework can be easily run on both platforms, thanks to the
kinematics/dynamics software abstraction layer presented in
\cite{frigerio16joser}.

\subsection{Outline}
%outline
The remainder of this paper is detailed as follows. In Section
\ref{sec:locFramework} we briefly present our statically stable gait
framework; Sections
\ref{sec:bodyMotionPhase} and \ref{sec:swingPhase}, detail the body motion and
swing motion phases of the gait, respectively; Section \ref{sec:swingPhase}
is particularly focused on the different stepping strategies,
depending on
the presence of a visual feedback; Section \ref{sec:reactiveFeatures} describes the
reactive modules for robust rough terrain locomotion; in Section
\ref{sec:terrainEstimation}, we present an
improved terrain estimator; Section \ref{sec:stairClimbing} shows a strategy for
stair climbing based on the proposed framework; in Section
\ref{sec:momentumBasedObserver}, we describe the implementation of our 
disturbance observer; in section \ref{sec:conclusions} we address the
conclusions.

\section{Locomotion Framework Overview}
\label{sec:locFramework}
In this section, we briefly illustrate
our \textit{statically} stable crawling framework,
(previously presented in \cite{focchi2017auro}),
enriched with additional features specific for rough terrain locomotion.
\begin{figure*}[tb]
\centering
\includegraphics[width=1.0\textwidth]{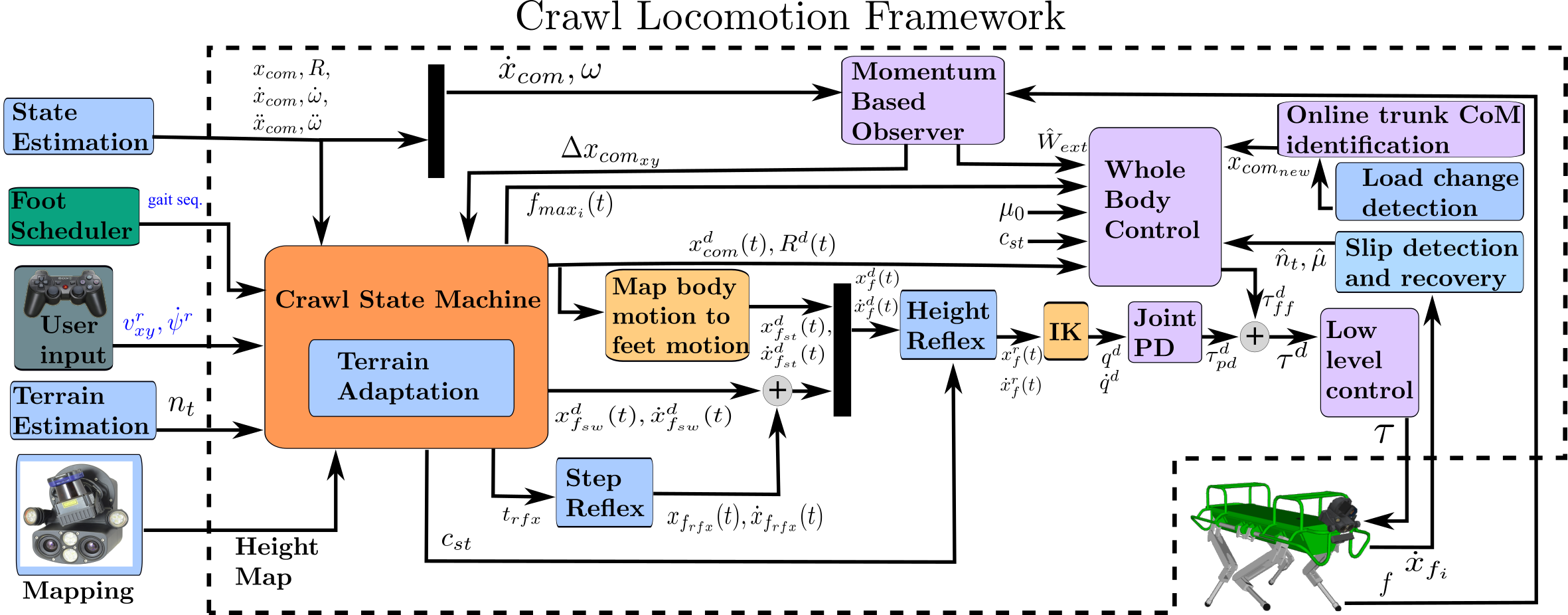}
\caption{Block diagram of the crawl  locomotion framework.
The crawl state machine (orange block) is further detailed
in Fig. \ref{fig:crawlStateMachine}. Please refer to $Appendix$ $B$ for a
description of the main symbols shown in this figure.}
\label{fig:blockDiagram}
\end{figure*}
Fig. \ref{fig:blockDiagram} shows the block diagram of the framework.
The core module is a state machine (see \cite{focchi2017auro} for details)
which orchestrates two temporized/event-driven locomotion phases:
the \textit{swing phase}, and the \textit{body motion phase}.
In the former, the robot has three legs in stance,
while in the latter it has four legs in stance.
%

%body motion phase
During the \textit{body motion phase}, the robot \gls*{com}
is shifted onto the \textit{future} support triangle,
(opposite to the \textit{next} swing leg,
in accordance to a user-defined foot sequence).
As default footstep sequence we use: RH, RF, LH, LF\footnote{LH,
LF, RH and RF stands for Left-Hind, Left-Front, Right-Hind and Righ-Front
legs, respectively.}.

The \gls*{com} trajectory is generated after the terrain
inclination is estimated (see Section \ref{sec:bodyMotionPhase}). At each
touchdown, the inclination is updated by fitting a plane through the
actual stance feet positions.
When the terrain is uneven (and the feet are not coplanar), an
\textit{average} plane is found.

%swing phase
The \textit{swing phase}
is a swing-over motion, followed by a linear searching motion (see
Section \ref{sec:swingPhase}) that terminates with a \textit{haptic}
touchdown.
The haptic touchdown ensures that the swing motion does not stop
in a prescheduled way; instead, the leg keeps extending until a new
touchdown is established.

%
%haptic touchdown
When a contact is detected (either by thresholding the \gls*{grf}, or
directly  via a foot contact sensor \cite{optoforce14patent, gao18patent}), the
touchdown is
established.
A searching motion with \textit{haptic} touchdown
is a key ingredient to achieve robust \textit{terrain
adaptation}.
Indeed, haptic touchdown is important to mitigate
the effect of tracking errors, because it allows to trigger the stance
only when the contact is truly stable.
%
%haptic feedback and vision
This is important whenever there is a
discrepancy between the plan and the real world.
This typically happens  when a vision feedback is
used to select the foothold (see Section \ref{sec:visionBasedStepping}),
because the accuracy of a height map
is typically in the order of centimeters \cite{mastalli18,
winkler15icra}.

%
%vision to estinate the normal to the terrain
The vision feedback can be also used to estimate the direction of the
normal of the terrain under the foot, in order to set the searching, or reaching, motion direction.
%The whole cycle duration (without the searching motion) $T_{body} + T_{sw}$
Both the body and the swing trajectories are generated as quintic polynomials.

\begin{figure}[bt]
\centering
\includegraphics[width=0.55\columnwidth]{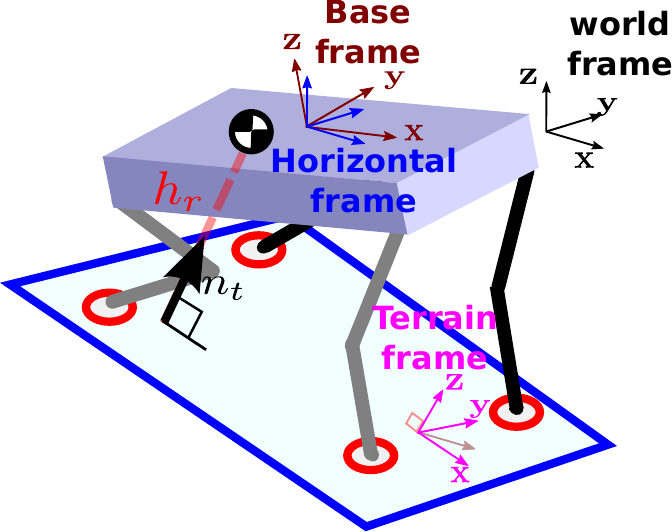}
\includegraphics[width=0.35\columnwidth]{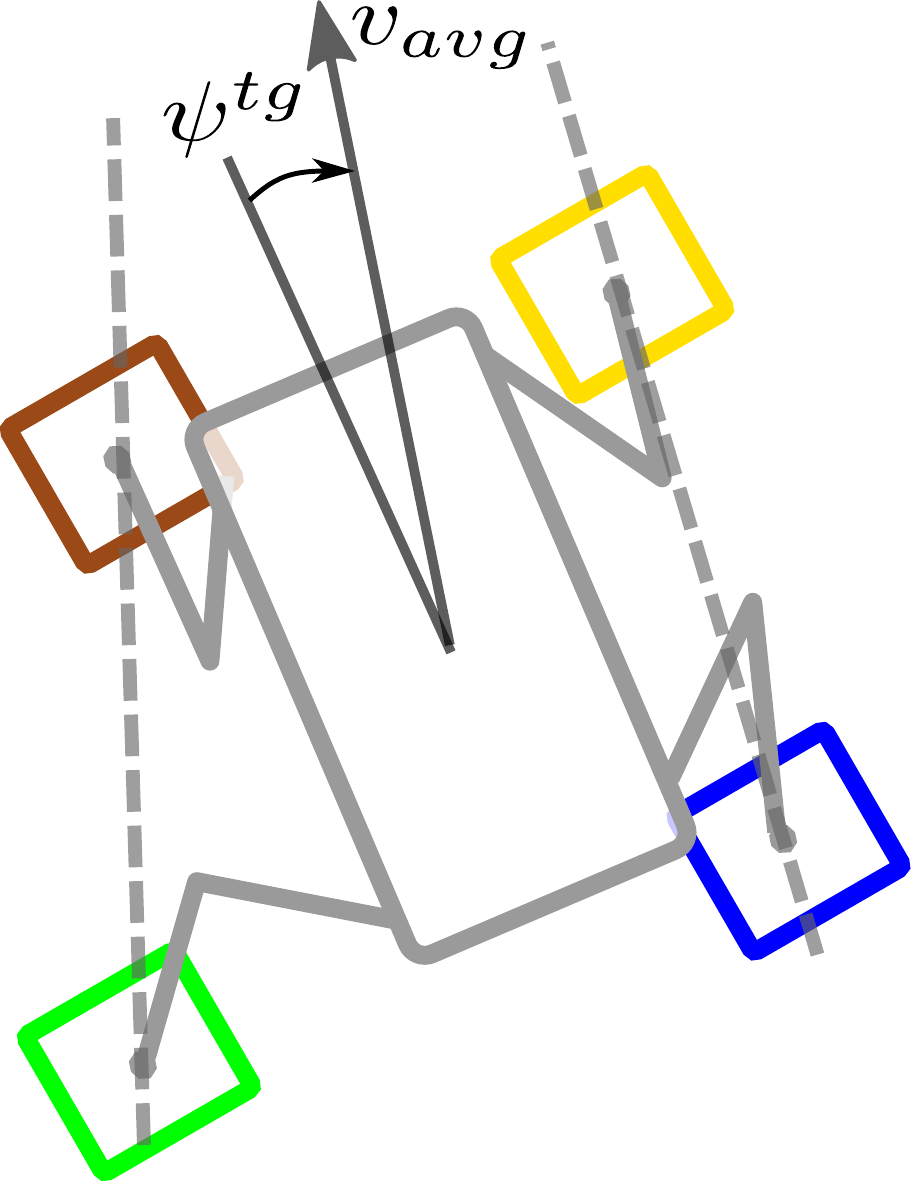}
\caption{(left) Definition of the relevant reference frames and of the terrain
plane (light blue) used in the locomotion framework and
(right) computation of the target $\psi^{tg}$ for the robot yaw.}
\label{fig:terrainFrame}
\end{figure}

The body trajectory is always expressed in the \textit{terrain frame}, which is
aligned
to the terrain plane (see Fig. \ref{fig:terrainFrame} (left)).
The terrain frame has the $Z$-axis
normal to the terrain plane, while the $X$-axis is a projection
of the $X$-axis of the base on the terrain plane,
and the $Y$-axis is chosen to form a right  hand side system of coordinates.
The swing trajectory is expressed in the \textit{swing frame}, which can be
either coincident
with the terrain frame (section \ref{sec:heuristicStepping})
or set independently for each foot (section \ref{sec:visionBasedStepping}),
depending on the stepping strategy adopted.
\begin{figure*}[tb]
	\centering
	\includegraphics[width=0.9\textwidth]{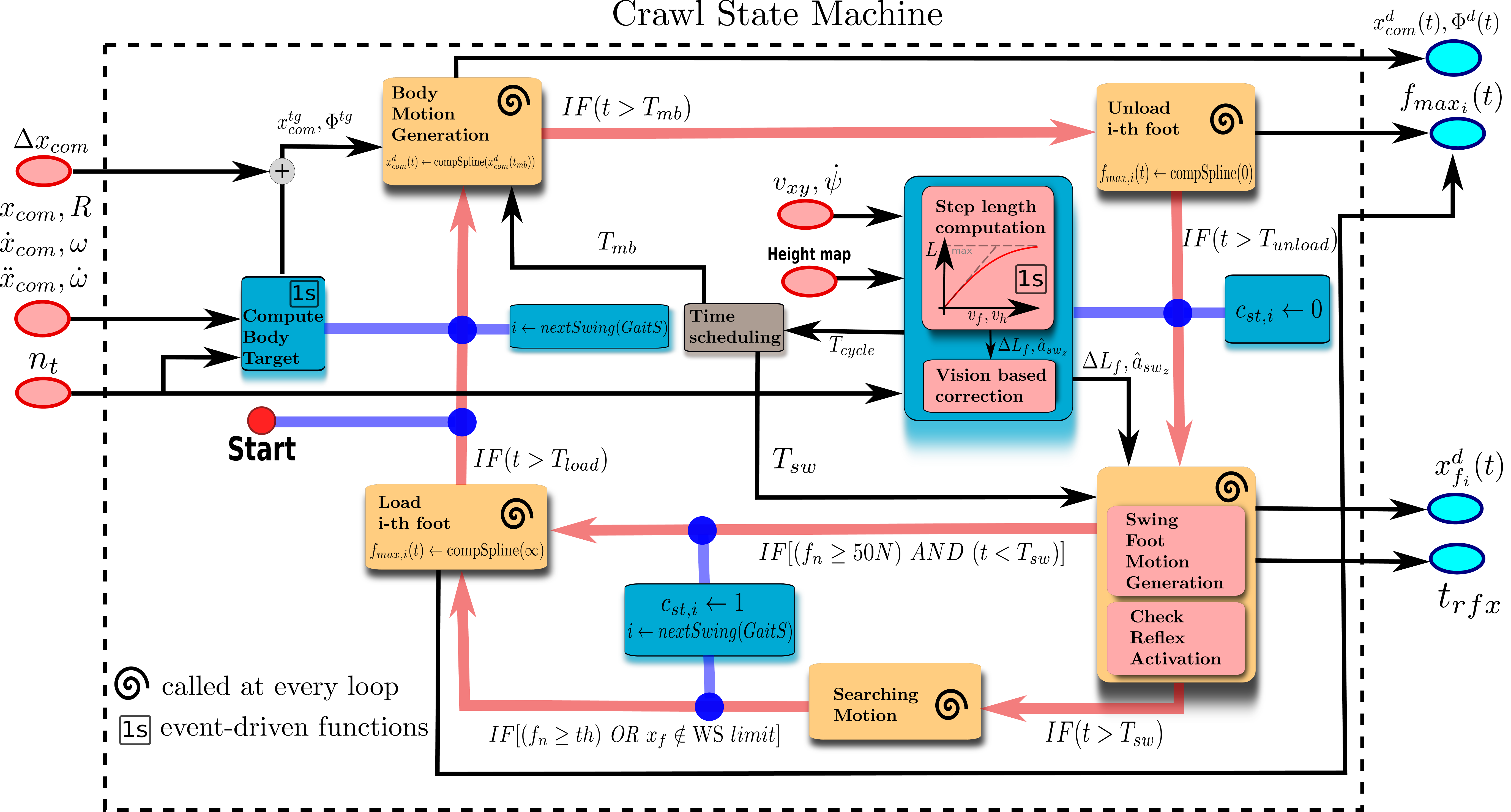}
	\caption{Logic diagram of the crawl state machine.
	The yellow rectangles represent the states, the red arrows represent the
	transitions while the blue boxes represent the actions associated to the
transitions.
	Blocks marked with 1s (1-shot) are event-driven and perform a single
	computation, while the other blocks
	(marked with the spiral) are called every loop. Please refer to
$Appendix$ $B$ for a short description of the main symbols shown in this figure.}
	\label{fig:crawlStateMachine}
\end{figure*}

%whole-body control
The \textit{Whole Body Control} module (also known as \textit{Trunk Controller}
\cite{focchi2017auro})
computes the torques required to control the position of 
the robot \gls*{com} and the orientation of the
trunk. At the same time, it redistributes the weight among the
stance legs ($c_{st}$) and ensures that friction constraints are not
violated.

%normal from vision improves whole-body control
The Trunk Controller action can be improved by setting
the real terrain normal (under each foot) obtained from a 3D map
\cite{mastalli18},
rather than using a default value for all the feet (\eg the normal to the
terrain plane).
This is particularly  important to avoid slippage when
the terrain shape differs significantly from a plane.

%impedance controller
To address unpredictable
events (\eg limit slippage on an unknown surface,  whenever the optimized force
is
out of the real (unknown) friction cone), we implemented an impedance
controller at the joint level \cite{boaventura15tro}\footnote{Without any
loss of generality, the same controller can be implemented at the foot level.
However, to avoid non collocation problems due to leg compliance,
it is safer to close the loop at the joint level.} in parallel  to the
whole-body controller.
The impedance controller computes the feedback joint torques $\tau_{fb} \in
\Rnum^n$
(where for \gls*{hyq} $n=12$ is the number of active joints) to
track reference joint trajectories $(q^d_j, ~\dot{q}^d_j  \in \Rnum^n)$.

The sum of the output of the Trunk Controller $\tau_{ff}^d \in \Rnum^n$ and of the impedance
controller $\tau_{pd}^d \in \Rnum^n$   form
the torque reference $\tau^d \in \Rnum^n$ for the
low level torque controller. If the model is accurate, the largest
term typically comes from the Trunk Controller.
%

%mapping from body motion to joint motion
Note that the  impedance controller should receive position and velocity
set-points that are consistent with the body motion, to prevent conflicts with
the Trunk Controller.
To achieve this, we map the \gls*{com} motion
into feet motion (see Section \ref{sec:bodyMotionPhase}) to provide
 the correspondent joint references~$q^d_j, ~\dot{q}^d_j$.

\section{Body Motion Phase}\label{sec:bodyMotionPhase}
%terrain adaptation
When traversing a rough terrain,
unstable footholds (\eg stones rolling under the feet),
tracking errors, slippage and estimation errors can create deviations from
the original motion plan.
These deviations require some corrective actions
in order to achieve \textit{terrain adaptation}.
In our framework, these actions are taken both at the beginning of
\textit{swing} and  \textit{body motion} phases.
The body motion phase starts with a foot touchdown and ends when the next foot 
in the sequence is lifted off the ground.
After each touchdown event, we compute the
target for the \gls*{com} position $x_{com}^{tg}$
and the body orientation  $\Phi^{tg}$ from
the \textit{actual}  robot state.
This feature prevents error accumulation and, together with the
haptic touchdown, constitutes
the core of the \textit{terrain adaptation} feature.

%trunk orientation
To avoid hitting kinematic limits, the robot's orientation should be
adapted to match the terrain shape. Indeed, some footholds can only be reached
by tilting the base. On the other hand, constraining the base to a given
orientation restricts the range of achievable motions.

%parametrization
We parametrize
the trajectory for the trunk orientation with Euler angles\footnote{This is a
reasonable choice because the robot is unlikely to reach singularity
(\ie \SI{90}{\degree} pitch).}
$\Phi^d(t) = [\phi(t)$, $\theta(t)$, $\psi(t)]$ (\ie roll,
pitch and
yaw respectively).
The trajectories are defined by quintic $3D$ polynomials connecting the
actual robot orientation (at the beginning of the phase,
namely the touchdown) $\Phi^d(0) =[\phi_{td}, \theta_{td},\psi_{td}]$
with the desired orientation $\Phi^d(T_{mb}) = \Phi^ {tg} =  [\phi^{tg},
\theta^{tg},\psi^{tg}]$,
where $T_{mb}$ is the duration of the \textit{move body phase}.
To match the inclination of the terrain, the target roll and pitch
are set equal to the orientation of the terrain plane that
was updated at the touchdown ($\phi^{tg} = \phi_t, \theta^{tg} = \theta_t$).
The target yaw $\psi^{tg}$ is computed to align the trunk with an average line
$v_{avg}$
from the ipsi-lateral\footnote{Belonging to the same side of the body.}
legs (see Fig. \ref{fig:terrainFrame} (right)).
%
%\begin{equation}
%\begin{array}{l}
%	v_l = {}_hR_b ({}_bx_{f_{LF}} - {}_bx_{f_{LH}})  \\
%	v_r = {}_hR_b ({}_bx_{f_{RF}} - {}_bx_{f_{RH}})\\
%	v_{avg} = (v_l + v_r)/2 \\  \psi^{tg} = atan2(v_{{avg}_y}, v_{{avg}_x});
%\end{array}
%\label{eq:yawShift}
%\end{equation}
%
%where ${}_hR_b \in SO(3)$ is the rotation matrix
%which maps vectors from the base frame $\mathcal{B}$
%to the \textit{horizontal frame} $\mathcal{H}$.
%
This is somewhat similar to a heading controller
that makes sure that the trunk ``follows'' the motion
of the feet (the feet motion is driven by the desired velocity from the user).
%

%com target
Similarly to the angular case, the \gls*{com} trajectory is initialized
with the \textit{actual} position of the \gls*{com}\footnote{We
found experimentally that using the \textit{desired} feet
position instead of the \textit{actual} one
to compute the \gls{com} target would
make the robot's height gradually decrease.},
while the target $x_{com}^{tg}$
 can be computed with different stability criteria
(\eg \gls*{zmp}-based \cite{Bretl2008} or wrench-based
\cite{orsolino18ral, Audren2017}).
Henceforth, the vectors are expressed in the world (fixed) frame
$\mathcal{W}$, unless otherwise specified.

\begin{figure}[tb]
	\centering
	\includegraphics[width=0.8\columnwidth]{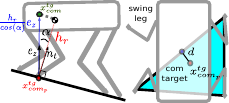}
	\caption{Heuristic generation of the \gls*{com} target. (left) planning
on	inclined terrain, (right) the future support polygon is depicted in
light blue 	while the projection $x_{{com}_p}^{tg}$ of the com target on the polygon
is a 	red dot.}
	\label{fig:planningStrategy}
\end{figure}

Since the crawl does not involve highly dynamic motions,
a heuristic \textit{static} stability criterion
can also be used and %  (that reduces the backward and lateral motions),
in particular, our strategy consists in making sure that the projection of the  \gls*{com} target
$x_{{com}_p}^{tg}$ always lies inside the future support triangle \cite{focchi2017auro}.
The robustness (in terms of stability margin) can be regulated by setting
the projected $CoM$ target at a distance $d$, on the support plane, from the
middle point of
the
segment connecting the diagonal feet (see Fig.~\ref{fig:planningStrategy}
(right)).
%
%robot height
We define the \textit{robot height} $h_r \in \Rnum$ as the distance between
the \gls*{com} and  the \textit{terrain plane} (see Fig.~\ref{fig:terrainFrame}
(left)).
This is computed
by averaging the actual positions
of the feet $_bx_{fi}$ in contact with the ground (stance feet):
\begin{equation}
h_r = e_z^T \frac{1}{c_{st}}\sum_{i=1}^{c_{st}} {}_tR_b ({}_bx_{f_i} +
{}_bx_{com})
\label{eq:height}
\end{equation}
where $c_{st}$ is the number of stance feet, ${}_bx_{com} \in \Rnum^3$ is the
\gls*{com} offset with respect to the base origin,
${}_tR_b \in SO(3)$ maps vectors from base to the
terrain frame and $e_z \in \Rnum^3$ selects the $Z$ component of $3D$ vectors.
In general, the robot height should be kept constant
(or it could vary with the cosine of the terrain pitch $\theta_t$ on ramps)
during locomotion.
If the above-mentioned heuristics is used for planning, the \gls*{com} target
$x_{com}^{tg}$
can be obtained by adding the height vector expressed in the world frame $h_r n_t$ (where $n_t$ is the
normal to the terrain) to
the
projection $x_{{com}_p}^{tg}$ coming from the heuristics.

%how to adjust  the height on inclined terrain
Note that, on an inclined terrain, to have the \gls*{com} above the desired
projection
and the distance of the \gls*{com} from the terrain plane
equal to $h_r$,  the following scaling should be applied (see
Fig.~\ref{fig:planningStrategy} (left)):
\begin{align}
&cos(\alpha) = e_z^T n_t \\
&x_{com}^{tg} = x_{{com}_p}^{tg} + \frac{h_r}{cos(\alpha)}e_z \nonumber
\label{eq:adjustHeight}
\end{align}
where $\alpha$ is the angle between the unit vectors $e_z$ and $n_t$.

Now, similarly to the orientation case,
we build a quintic polynomial $x_{com}^d(t)$
to connect the \textit{actual} \gls*{com} at
the beginning of the motion phase
(namely at the touchdown) $x_{com,td}$ to the computed target $x_{com}^{tg}$.

To determine the 6 parameters of each quintic,
in addition to the initial/final positions,
we force the initial/final velocities and accelerations to zero,
both for \gls*{com} and Euler angles.
This ensures \textit{static stability}\footnote{Statically stable gaits
are convenient for locomotion in dangerous environments (\eg nuclear
decommissioning missions) because the motion can be stopped at any time.
However, without loss of generality, the final velocity can be set to
any other arbitrary value other than zero.}
and implies that the robot's trunk does not move when
one leg is lifted from the ground (\ie during the swing phase).

%wrench based optimization
Alternatively, if a wrench-based optimization is used, as in
\cite{orsolino18ral}, the robot height
should be constrained  (\eg to be constant)
and the \gls*{com} $X,Y$ trajectory will be a result of the optimization.

%mapping to feet motion
As mentioned in the previous section,
it is convenient to provide desired joint positions (for the stance legs)
that are consistent with the body motion.
We first map the body motion into feet motion.
This mapping is linear in the velocity
domain and can be computed independently for each stance foot $i$ as:
\begin{equation}
\dot{x}_{f_i}^d[k] = - \dot{x}_{com}^d[k] - \omega[k] \times
\left(x_{f_i}^d[k-1] - x_{com}([k]\right)
\label{eq:mapping}
\end{equation}
where the desired position of the foot $x_{f_i}^d[k-1]$ at the previous controller loop is
used.
Then,  $x_{f_i}^d[k]$ is obtained by integrating the velocity $\dot{x}_{f_i}^d[k]$ (\eg with a trapezoidal rule).
%Note that, since we design a trajectory for the \gls*{com} we consider
% $x_{b,com} \in \Rnum^3$ in \ref{eq:mappingt}
%that is the offset of the \gls*{com} from the origin of the base frame.
Afterwards, we compute the corresponding desired joint positions $q_i \in \Rnum^3$
through inverse kinematics: $ q_i = IK(x_{f_i})$  and the joint
velocities as:
\begin{equation}
 \dot{q}^d_i = J(q_i)^{-1} \dot{x}_{f_i}
\end{equation}
where $J_i \in \Rnum^{3\times3}$ is  the Jacobian of foot $i$\footnote{Note that
we performed a simple inversion since in our robot we have point feet and 3
\gls*{dof} per leg, thus the Jacobian matrix is squared.}.
\section{Swing phase}\label{sec:swingPhase}
The swing phase of a leg starts with the foot liftoff and ends with the foot
touchdown. The obvious goal of the leg's swing phase is to establish a new
foothold.
Then, an \textit{interaction force} drives the robot's trunk towards the desired
direction.
The swing phase has two main objectives: 1) attaining enough clearance to
overcome potential obstacles (so as to avoid \textit{stumbling}) and
2) achieving a stable
contact.
\subsection{Heuristic Stepping}
\label{sec:heuristicStepping}
In this section, we illustrate the heuristic \textit{stepping strategy}  we use
for \textit{blind} locomotion.
The goal is to select the footholds to track a \textit{desired} speed from the
user when no map of the surroundings is available.

First, we compute the default step length (for the swing leg)
from the desired linear  and heading  velocities $v_{xy}^{r} \in \Rnum^2,
\;\dot{\psi}^{r} \in
\Rnum$.
For sake of simplicity, \textit{only in this
section} the vectors  are expressed in the \textit{horizontal frame} $\mathcal{H}$ 
\footnote{The horizontal frame  $\mathcal{H}$ is the reference frame
	that shares the same  origin and yaw  value with the base frame but
	is aligned (in pitch and roll) to the world frame,
	hence horizontal (see Fig \ref{fig:terrainFrame}~(left)).} (instead of the \textit{world frame} $\mathcal{W}$), unless
otherwise specified. Expressing the default step in a \textit{horizontal frame}
allows to formulate the swing motion independently
from the orientation of both terrain and trunk.

\begin{figure}[bt]
\centering
\includegraphics[width=0.6\columnwidth]{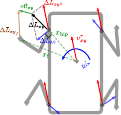}
\caption{Vector definitions for the heuristic stepping strategy. 
	Red arrows represent the desired linear velocity  $v_{xy}^{r} $ 
	and the linear component $\Delta L_{xy0}$ of the default step. 
	In blue is the desired heading velocity $\dot{\psi}^{r}$ 
	and the angular component $\Delta L_{{h0}}$
	of the default step $\Delta \bar{L}_{xy0}$ that is in black. 
	The offset to increase/decrease the stance size is in green, 
	The step about the foot $\Delta L_{xy}$ is depicted in brown.}
\label{fig:steppingStrategy}
\end{figure}

The swing trajectory consists of a parametric curve, whose four main parameters
are: the linear foot displacement $\Delta L_{x0},\Delta L_{y0}$, the angular
foot displacement $\Delta H_0$ (see Fig.
\ref{fig:steppingStrategy}), and the default swing duration $T_{sw}$. These
quantities can be computed from the nonlinear mapping $F(\cdot)$:
\begin{equation}
\mat{\Delta L_{x0} & \Delta L_{y0} & \Delta H_0 &  T_{sw}} = F(v_{xy}^{r},
\dot{\psi}^{r})
\label{eq:defaultStep}
\end{equation}
%
%For sake of simplicity, the heading velocity is defined about
%the base origin rather than at the \gls*{com}.

$F(\cdot)$ makes sure that the step lenght increments linearly with the
desired velocity, but only at low speeds (see \textit{compute step length}
block in Fig.
\ref{fig:crawlStateMachine}).
When the step length approaches the maximum value,
the cycle time $T_{cycle}$ (sum of swing and stance time of each leg) is
decreased, and the stepping frequency  $f_s = 1/T_{cycle}$ is increased
accordingly to avoid hitting the kinematic limits.
For instance, for the $X$ component, we can use the following equation:
\begin{align}
A &=  \frac{2 \Delta L_{x0}^{max}}{\pi}  \\
G &= \frac{ \Delta L_{x0}^{tr}}{\Delta L_{x0}^{max}v_{tr}} \\
\Delta L_{x0} &=A \cdot \text{arctan}(G v_{x}^{r})
\label{eq:velMapping}
\end{align}
where $ \Delta L_{x0}^{max}$ is the maximum
allowable step length in the $X$ direction.

According to \eref{eq:velMapping}, $\Delta L_{x0}$ linearly
increases with velocity up to the transition point $\Delta L_{x0}^{tr}$,
which corresponds to the user-defined value $v_{tr}$.
Then, the step length is increased sub-linearly with velocity (because
the stepping frequency is also increased), up  to the saturation point $\Delta
L_{x0}^{max}$. After this point, only the frequency increases.
Similar computations are performed for  $ \Delta L_{y0}$ and  $ \Delta H_{0}$.
Since it is possible to set different values for heading and linear speed, the
cycle time (and so the stepping frequency)
is adjusted to the minimum coming from the three velocity components:

\begin{equation}
T_{cycle} = min\left(\frac{\Delta L_{x0}}{v_{x}^{r}}, \quad \frac{\Delta
L_{y0}}{v_{y}^{r}}, \quad  \frac{\Delta H_{0}}{\dot{\psi}^{r}}\right)
\label{eq:defaultStep2}
\end{equation}

%if some velocity term goes below a certain threshold, the contribution
%of that component to the cycle time $T_{cycle}$ is set very big.
When a new value of $T_{cycle}$ is computed, the durations of the body and swing
trajectories are recomputed as $T_{mb} = (T_{cycle} -~ T_{lu})D$ and
 $T_{sw} = (T_{cycle} - T_{lu})(1-D)$ respectively, according to the
\textit{duty factor} $D$ \cite{alexander2003principles}.

The variable $T_{lu}$ represents the cumulative duration of the load/unload
phases.
As explained in \cite{focchi2017auro}, we remark that a load/unload phase
at the touchdown/liftoff events
is  important to avoid torque discontinuities.
In particular, a \textit{load phase} at the touchdown
allows the Trunk Controller to redistribute smoothly the load on all the legs.
At the same time, it ensures that the \gls*{grf}
always stay inside the friction cones, thus reducing the possibility of
slippage.

Note that a heading displacement $\Delta H_0$ can be converted into a
$X,Y$ displacement of the foot, as shown in Fig. \ref{fig:steppingStrategy},
by:
\begin{equation}
\Delta L_{h0} =  E_{xy} \mat{0& 0& \Delta H_0}^T\times x_{\text{hip}}
\label{eq:headingCorrection}
\end{equation}
where $x_{\text{hip}} \in \Rnum^3$ is the vector from the origin of the base
frame to the hip  of the swinging leg,  and $E_{xy} \in \Rnum^{2 \times 3}$
selects the $X,Y$
components.
Then, the default step becomes:
\begin{align}
\Delta \bar{L}_{xy0} = \Delta L_{xy0} + \Delta L_{{h0}}.
\label{eq:stepxy}
\end{align}

Note  that we defined  the \textit{default}
step  about the hip rather than
the  foot position (see Fig. \ref{fig:steppingStrategy}). This is crucial  to
avoid inconvenient kinematic configurations while walking (\eg stretched
or ``crouched'' configuration), thus degenerating the support polygon.
Indeed, if we refer each step to the previous foot position,
an anticipated/delayed touchdown would produce unexpected
step lengths. This would make the stance feet closer/farther over
time\footnote{As a matter of fact, if the support polygon shrinks, the
robustness decreases. In particular, \gls*{com} tracking errors and external
pushes can move the \gls*{zmp} very close to the boundary of the support
polygon.
This would result in a situation where not all the contact feet are
``pushing'' on the ground
(\eg the robot starts tipping about the line connecting two feet).}.

Since the swing polynomial is defined at the foot
level, we have to determine the step
$\Delta L_{xy}\in \Rnum^2$  from $\Delta \bar{L}_{xy0}$, to express it about
the actual foot position:
\begin{align}\label{eq:stepxy_foot}
\Delta L_{xy} = \Delta \bar{L}_{xy0} + E_{xy} (\text{off}_{xy} + x_{hip} -
x_{f}),
\end{align}
where $\text{off}_{xy}$ is a parameter to adjust the average size of the stance
polygon.

In delicate situations, it  might be useful
to walk with the feet more outward to increase locomotion
stability at the price of a bigger demanded torque at the \gls*{HAA} joint.

\begin{figure}[tb]
\centering
\includegraphics[width=1.0\columnwidth]{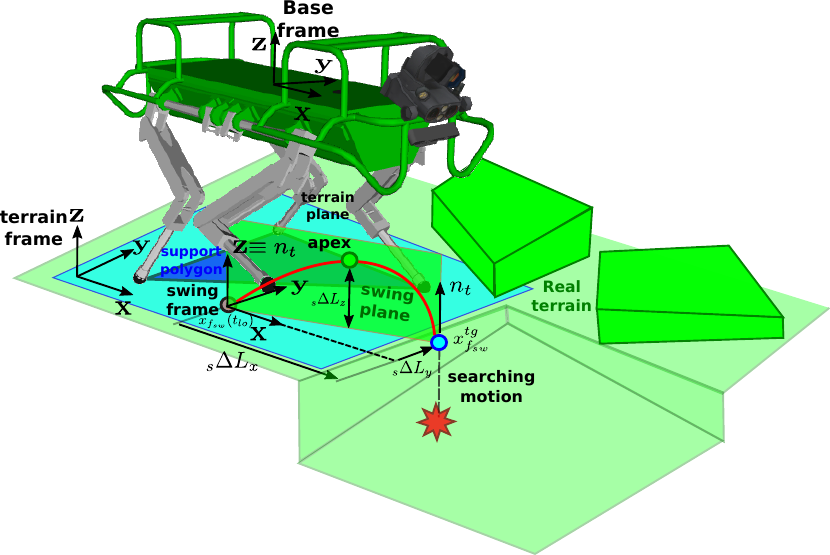}
\caption{Swing frame and terrain frame definition for the heuristic-based
stepping strategy used in blind locomotion.
The terrain presents a decrease in elevation, 
 haptically handled by the searching motion.}
\label{fig:swingFrameHeuristic}
\end{figure}

As a final step, it is convenient to express the swing motion
in the \textit{swing plane} (see Fig. \ref{fig:swingFrameHeuristic})\footnote{We
	call the \textit{swing plane} the plane passing though the X-Z axes of
the swing
	frame.}. This means 
we need to express the above quantities in the 
swing frame $\mathcal{S}$\footnote{The \textit{swing frame}, in general, is
aligned with
the \textit{terrain frame} unless a vision based stepping strategy is used.
In this case, the swing frame is  computed \textit{independently} for each foot
(as explained in Section \ref{sec:visionBasedStepping}) from the visual input.}.

We set quintic $3D$ polynomials $p(t) \in \Rnum^3$ of duration
$T_{sw}$,
where the $X$, $Y$ components go from $(0,0)$ to  ${}_s\Delta L_{xy}$,
while for the $Z$ component we set two polynomials
such that the swing trajectory passes by an intermediate \textit{apex}
point at a time $T_{sw} \chi$.
At the apex, the $Z$ component is equal to the step height ${}_s\Delta L_{z}$
and $\chi \in [0,1]$ represents the apex ratio that can
be adjusted to change the apex location (see Fig.
\ref{fig:swingFrameHeuristic}).
Hence, the swing foot reference trajectory is computed from the
\textit{actual} foot position at the instant of liftoff
to the desired foothold as:
\begin{align}
x_{f_{sw}}^d(\bar{t}) =  x_{f_{sw,lo}} + p(\bar{t}),
\label{eq:stepxy_foot2}
\end{align}
where $\bar{t} \in \mat{0& T_{sw}}$, $T_{sw}$ is the swing duration, and
the final target is defined as $x_{f_{sw}}^{tg} = x_{f_{sw,lo}} +
p(T_{sw})$.

\textbf{Remark}: the \textit{swing frame} (unless specified) is always aligned
with the \textit{terrain frame}.
Consequently, we have an initial \textit{retraction}
along the normal to the \textit{terrain plane} (thus avoiding
possible trapping or stumbling of the foot),
while the step is performed \textit{along} the terrain plane.
A swing motion expressed in this way allows to
adjust 
the step clearance simply with the step height ${}_s\Delta L_{z}$. This
parameter  regulates the maximum retraction from the terrain (apex point).
In general, the apex is located in the middle
of the swing, but it can be parametrized to shape differently 
the swing trajectory.

\textbf{Definition}: \textit{Searching motion}.
A \textit{haptic} triggering of the touchdown allows to accommodate the shape of
the
terrain by stopping the swing motion either
before of after the foot reaches the target $x_{f_{sw}}^{tg}$.

If the touchdown is deemed to happen after the target is reached, the
trajectory is continued
with a leg extension movement that we name \textit{searching motion}: the swing foot keeps
moving linearly along the
direction of the terrain plane normal $n_t$ (see Fig.
\ref{fig:swingFrameHeuristic})
until it  touches the ground or eventually reaches the workspace (WS) limit.
In this way, the stance is triggered in any case.
To avoid  mobility loss, a height reflex can be enabled
to aid the \textit{search} motion with the
other stance legs (see section \ref{sec:heightReflex}).
\subsection{Vision Based Stepping}
\label{sec:visionBasedStepping}
\begin{figure}[htb]\label{fig:swingFrameVision}
\centering
\includegraphics[width=1.0\columnwidth]{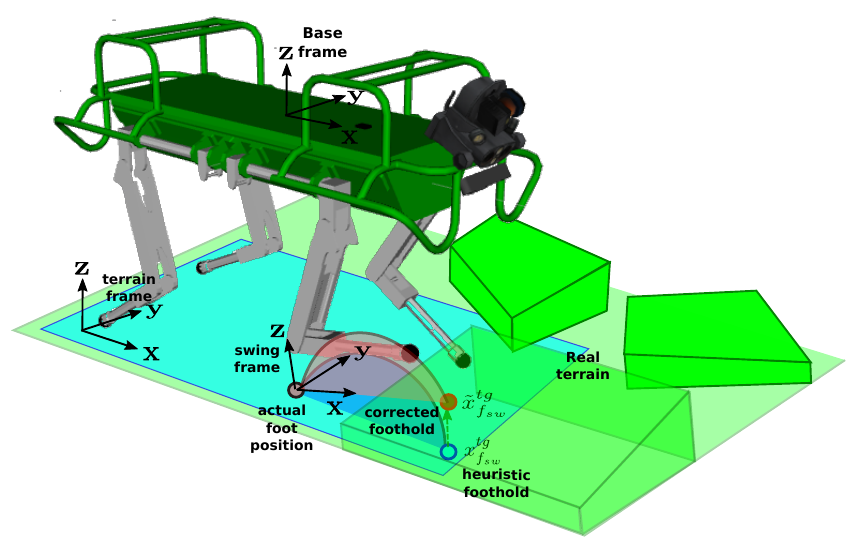}
\caption{Swing frame definition for the vision-based stepping strategy.
Blue and red shaded area represent the swing
in the heuristic and the vision-based cases, respectively.
In this case the terrain elevation is higher
than the foot location at the liftoff and the target
foot-hold (blue dot) computed with heuristics,
is corrected by vision (red dot) to lay on the real terrain.
The swing frame is also adjusted accordingly. }
\end{figure}

The \textit{terrain plane} is a very coarse approximation of the terrain.
If a vision feedback is available, it is
advisable
to exploit this information, which allows to:
\begin{enumerate}
\item compute the target foothold on the \textit{actual} terrain
rather than on its planar approximation (see Fig. \ref{fig:swingFrameVision}).
This allows to increase the overall swing \textit{clearance} (cf. blind
stepping in Fig.
\ref{fig:swingFrameHeuristic} with vision based
stepping in Fig. \ref{fig:swingFrameVision}).
\item set a different swing frame for each
foot instead of having the \textit{swing frame} always coincident to the
\textit{terrain frame}. This enables the swing on different planes per each
leg (essential for climbing though difficult terrain configurations such as
 V-shaped walls \cite{focchi2017auro}).
\end{enumerate}

To implement the first point, we first compute a  step (along the terrain plane)
using the heuristic-based stepping strategy. This is meant to realize the
user velocity. Then, the  height map of
the terrain is queried at the target location, and the $Z$ component of the
target is
corrected to have it on the real terrain ($X,Y$ components remain unchanged):
\begin{equation}
\tilde{x}^{tg}_z = H\left(E_{xy}x^{tg}_{f_{sw}}\right)
\end{equation}
where $E_{xy} \in \Rnum^{2 \times 3}$ selects the  X,Y components, and
$H(\cdot)$
is a function that queries   the terrain elevation for a certain (X,Y)
location
on the X-Y plane\footnote{Being the map expressed in a (fixed)  world frame, it is necessary to apply appropriate kinematic
conversions to this frame before evaluating the map.}.
%
%\begin{align}
%&\tilde{x}^{tg}_{f_{sw,xy}} = E_{xy} ({}_h{x}^{tg}_{f_{sw}}  + x_b) \\
%&\tilde{x}^{tg}_{f_{sw,z}} = H( \tilde{x}^{tg}_{f_{sw,xy}}) \nonumber
%\label{eq:newTargetw}
%\end{align}
%
%where $x_b$ is the location of the origins both  of the horizontal and of the
%base frames,
%and $E_{xy} \in \Rnum^{2 \times 3}$ selects the  X,Y components, and $H(\cdot)$
%is a function that queries height  of the terrain  for a certain (X,Y) location
%in the X-Y plane.
%\footnote{The world frame and the horizontal frame are  aligned so it is licit
%to add ${}_h{x}^{tg}_{f_{sw}}$
%with $x_b$}

\begin{figure}[htb]
\centering
\includegraphics[width=0.7\columnwidth]{%
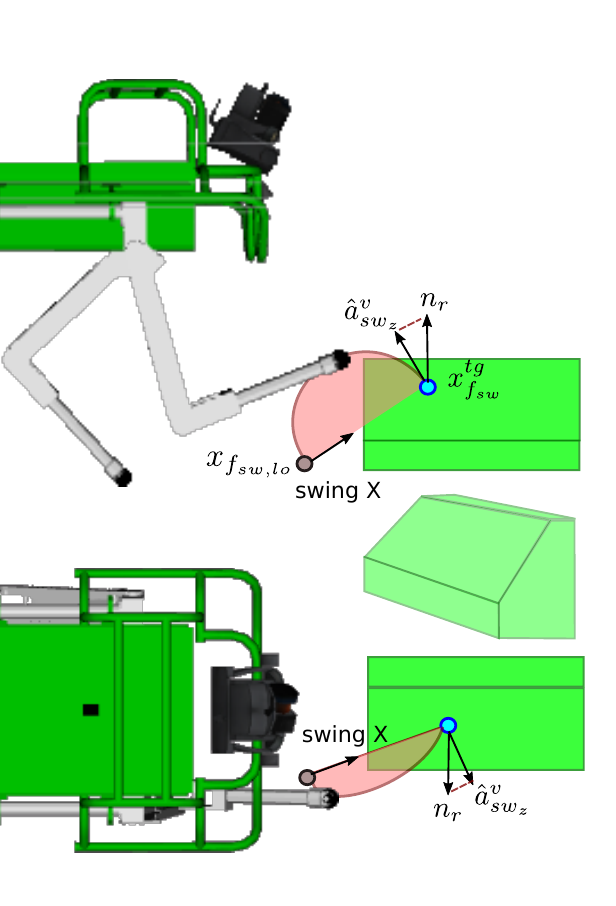}
\caption{Vision based adjustment of the target foothold Z coordinate. Top and lateral views of the swing
plane on a slanted terrain.}
\label{fig:swingFrameAdjustmentVision}
\end{figure}

As a final step, we  compute the $Z$ axis $\hat{a}_{{sw}_z}$ of the swing frame
such that its $X$ axis $\hat{a}_{{sw}_x}$ is
aligned  with the segment connecting the actual foot position to the
corrected foothold $\tilde{x}_{f_{sw}}^{tg}$ (see Fig
\ref{fig:swingFrameAdjustmentVision}):

\begin{align}
\label{eq:swingFrameVision}
&a_{{sw}_z} =  (\tilde{x}^{tg}_{f_{sw}} -  x_{f_{sw}} )  \times  ({}_wR_b e_y),
\nonumber\\
& \hat{a}_{{sw}_z} =  \frac{a_{{sw}_z}}{\Vert a_{{sw}_z}  \Vert},  \\
&\hat{a}_{{sw}_y} = \hat{a}_{{sw}_z} \times e_x,  \nonumber \\
&\hat{a}_{{sw}_x} = \hat{a}_{{sw}_y} \times \hat{a}_{{sw}_z}, \nonumber
\end{align}
where the $\hat{(\cdot)}$ represents unit vectors and $e_x, e_y$
are the base frame axes expressed in the world frame.
This correction allows for more clearance
around the actual terrain during the swing motion
and reduces the chances of stumbling.
%and reduces the searching motion
%(e.g. if $\tilde{x}^{tg}_{f_{sw}}$ is not on the terrain)
%because the trajectory is tailored for a specific target point.

The first consequence is that the swing
frame  is no longer aligned with the terrain frame,
and the swing trajectory is laying on the plane
$\hat{a}_{{sw}_x}/\hat{a}_{{sw}_z}$.

Additionally, if  the
normal $n_r$  of the \textit{actual}  terrain is available
from the visual feedback \cite{mastalli18},
we can exploit it to further correct the swing plane,
in order to have the swing-down tangent to that normal.
This is particularly useful when the robot has to step
on a laterally slanted surface (e.g. like in \cite{focchi2017auro}).

In our experience, however, we noticed that on the sagittal direction, maximizing
clearance has more priority. Therefore,  we remove the component of $n_r$
along the $X$ axis computed in \eref{eq:swingFrameVision} ($\hat{a}_{{sw}_z}^v
= n_r -
\hat{a}_{{sw}_x}^Tn_r)\hat{a}_{{sw}_x}$) and use as the new swing $Z$ axis.
The new vision-corrected  Z-axis of the
\textit{swing frame} becomes $n_{rp}$ (see Fig.
\ref{fig:swingFrameAdjustmentVision}),
while the other axes should be recomputed accordingly as in
\eref{eq:swingFrameVision}.
\subsection{Clearance Optimization}
\label{sec:clearanceOptimization}
\begin{figure}[tb]
	\centering

\includegraphics[width=0.7\columnwidth]{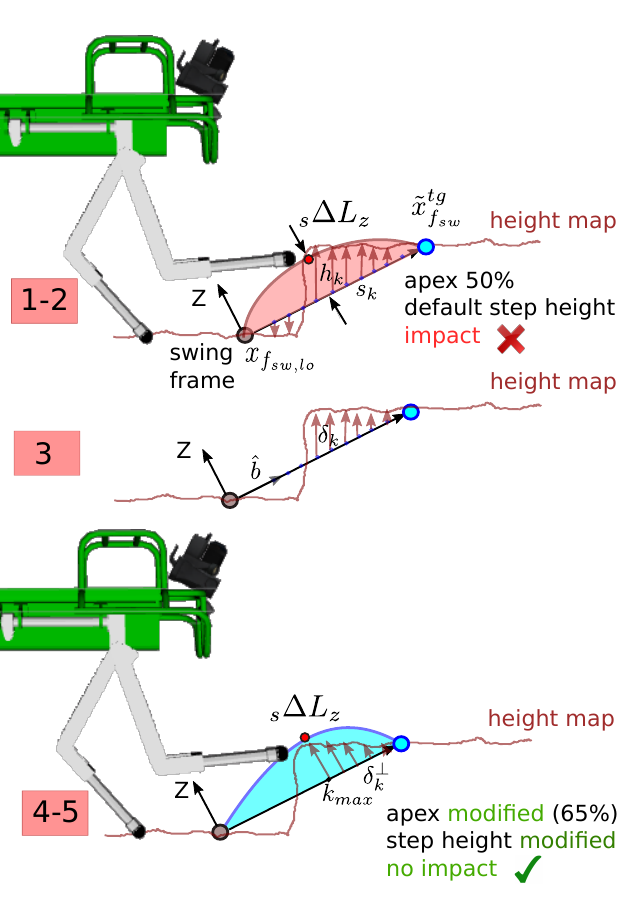}
	\caption{Clearance optimization. The grey dot represents the actual foot
location at liftoff, the blue dot the foot target while
    the red dot the apex location. The numbers in the red rectangles are related
to the steps in \eref{eq:clearanceOptimSteps}.}
	\label{fig:clearanceOptimization}
\end{figure}

If the quality of the map is good enough, the
\textit{apex}
(point of maximum clearance) and the step height ${}_s\Delta L_{z}$
can be adjusted in accordance with the point of maximum asperity of the
terrain, so as to maximize clearance. To some extent, this
can be considered a simple adaptation of the swing  to the
terrain shape.
First, we take a slice of the map and evaluate the  height of the terrain
on the line segment of direction $\hat{b}$ connecting the actual foot location with the target $\tilde{x}^{tg}_{f_{sw}}$
 (see Fig. \ref{fig:clearanceOptimization}).
With a discretization of $N$ points, we then follow the steps below:
\begin{align}
\label{eq:clearanceOptimSteps}
1)\,\, &s_k = \tilde{x}^{tg}_{f_{sw}} \frac{k}{N} + \tilde{x}_{f_i} \left(1 -
\frac{k}{N}\right),  \quad k=1...N,  \nonumber \\
2)\,\, &h_k = H(E_{xy}s_k),  \nonumber \\
3)\,\, &\delta_k =  \mathcal{P_{+}} (h_k - e_z^T s_k),  \\
4)\,\, &\delta_k^{\perp} =  \left\Vert  \mat{0&0& \delta_k}^T -
\left(\mat{0&0&\delta_k} \hat{b} \right) \hat{b} \right\Vert, \nonumber\\
5)\,\, &\mat{{}_s\Delta L_{z} & k_{max}}  = \max_{k}  \delta_k^{\perp},
\nonumber
\end{align}
where 1) is a discretization of the line segment; 2) is the evaluation of the
height map on the discretized points;
3)  $\mathcal{P_{+}}(\cdot):\Rnum \rightarrow \Rnum$ is a function that sets
negatives values to zero;
in 4) we project the $\delta_k$ onto the swing $Z$ axis, getting
$\delta_k^{\perp}$; in 5) we set the step height ${}_s\Delta L_{f_{sw}}$ as the
maximum value among
the
$\delta_k^{\perp}$ where $k_{max}$ is its index.
Finally, the apex can be set to $k_{max}/N$.
This reshapes the swing in a conservative way,
by adjusting the apex according to the terrain feature.

\subsection{Time Rescheduling}
\label{sec:timeRescheduling}

In our state-machine-based framework, a change in locomotion speed
corresponds to changing the duration of the swing/body phases.
However, to change speed promptly (without waiting until the end of the
current phase)
it is necessary to reschedule the polynomials of the active phase (swing or body
motion). As a consequence, we compute
a new duration $T_f^{'} $ (where $T_f$ is the previous duration).
Setting a new duration for a (previously designed) polynomial is equivalent to finding the \textit{initial point} of a new polynomial
that: 1) has the new duration $T_f^{'}$ and 2) is passing
through the same  point at the moment of the rescheduling.
We know that a quintic polynomial can be expressed as:
\begin{equation}
p(t) = a^T \mu(t), \quad \dot{p}(t) = a^T \dot{\mu}(t), \quad \ddot{p}(t) = a^T
\ddot{\mu}(t)
\label{poly}
\end{equation}

where:
\begin{align}\label{eq:polycoeff}
&a^T = \mat{a_5 & a_4 & a_3 & a_2 & a_1 & a_0}, \\
&\mu(t) = \mat{t^5 & t^4 & t^3 & t^2 & t & 1}, \nonumber \\
&\dot{\mu}(t) = \mat{5t^4 & 4t^3 & 3t^2 & 2t & 1 & 0}, \nonumber \\
&\ddot{\mu}(t) = \mat{20t^3 & 12t^2 & 6t & 2 & 0 & 0}. \nonumber
\end{align}

Setting the initial/final boundary conditions for position, velocity and
acceleration:
$p_0 = a^T\mu(0)$, $\dot{p}_0 = a^T\dot{\mu}(0)$, $\ddot{p}_0 =
a^T\ddot{\mu}(0)$,
$p_f  = a^T\mu(T_f)$, $\dot{p}_f = a^T\dot{\mu}(T_f)$, $\ddot{p}_f =
a^T\ddot{\mu}(T_f)$
is equivalent to solving a linear system
of $6$ equations that allows us to find the $a_i$ parameters:
\begin{small}
\begin{align}\label{eq:polycoeff2}
   & a_0 = p_0,\\
   & a_1 = \dot{p}_0, \nonumber\\
   & a_2 = 0.5 \ddot{p}_0, \nonumber\\
   & a_3 = \frac{1}{2T_f^3}\left[20p_f -20p_0 + T_f(-12\dot{p}_0  -
8\dot{p}_f) + T_f^2(-3 \ddot{p}_0 + \ddot{p}_f)\right], \nonumber\\
   & a_4 =\frac{1}{2T_f^4} \left[30p_0 - 30p_f + T_f(16\dot{p}_0 + 14\dot{p}_f)
+ T_f^2(3\dot{p}_0 - 2\dot{p}_f)\right],  \nonumber\\
   & a_5 = \frac{1}{2T_f^5} \left[12p_f -12p_0 + T_f(-6\dot{p}_0 -6\dot{p}_f)
+ T_f^2(-\ddot{p}_0 + \ddot{p}_f)\right],  \nonumber
\end{align}
\end{small}
with a similar criterion, to ensure continuity in the position,
we can compute the new initial point $p_0^{\prime}$ such that:
\begin{equation}
a^{\prime}\mu(\bar{t}) = p(\bar{t}),
\label{swingreplanning}
\end{equation}
where $\bar{t}$  is the time elapsed (from $0$) at the moment of the
rescheduling,
and  $a^{\prime}$ are the coefficient of the new polynomial of duration
$T_f^{'}$.
Then, collecting $p_0$ from all the parameters in \eref{eq:polycoeff2},
we can obtain a closed form  expression for it:

%\setlength{\medmuskip}{2\medmuskip}% Formerly 4.0mu plus 2.0mu minus 4.0mu ->
% 4.0mu
%\setlength{\thickmuskip}{2\thickmuskip}% Formerly 5.0mu plus 5.0mu -> 5.0mu
%\setlength{\thinmuskip}{2\thinmuskip}% Formerly 5.0mu plus 5.0mu -> 5.0mu
\begin{small}
\begin{align}
 \label{eq:initialPoint}
   & \beta_{1} = \frac{1}{2T_f^3}\left[ 20p_f + T_f(-12\dot{p}_0  - 8\dot{p}_f)
+ T_f^2(-3 \ddot{p}_0 + \ddot{p}_f)\right],  \nonumber\\
   & \beta_{2} = \frac{1}{2T_f^4}\left[- 30p_f + T_f(16\dot{p}_0 + 14\dot{p}_f)
+ T_f^2(3\dot{p}_0 - 2\dot{p}_f)\right],   \nonumber\\
  & \beta_{3} = \frac{1}{2T_f^5}\left[ 12p_f + T_f(-6\dot{p}_0 -6\dot{p}_f) +
T_f^2(-\ddot{p}_0 + \ddot{p}_f)\right],   \nonumber\\
  & \beta_{4} = 1  -10T_f^3\bar{t}^3 + 15T_f^4\bar{t}^4 - 6T_f^5\bar{t}^5,
\nonumber
\end{align}
\vspace{-.5cm}
\begin{align}
  p_0 &= \frac{1}{\beta_{4}} \left[ p(\bar{t}) - a_1\bar{t} + a_2\bar{t}^2 +
\beta_{1}\bar{t}^3 + \beta_{2}\bar{t}^4 + \beta_{3}\bar{t}^5 \right]
\end{align}
\end{small}

Now, exploiting \eref{eq:initialPoint} for the computation of $p_0$, the new polynomial parameters can be recomputed as in \eref{eq:polycoeff}
while  keeping the other boundary conditions unchanged.
This will result in a polynomial that \textit{continues} from
$\bar{t}$ with a new duration $T_f^{'}$.

\begin{figure}[tb]
\centering
\includegraphics[width=1.0\columnwidth]{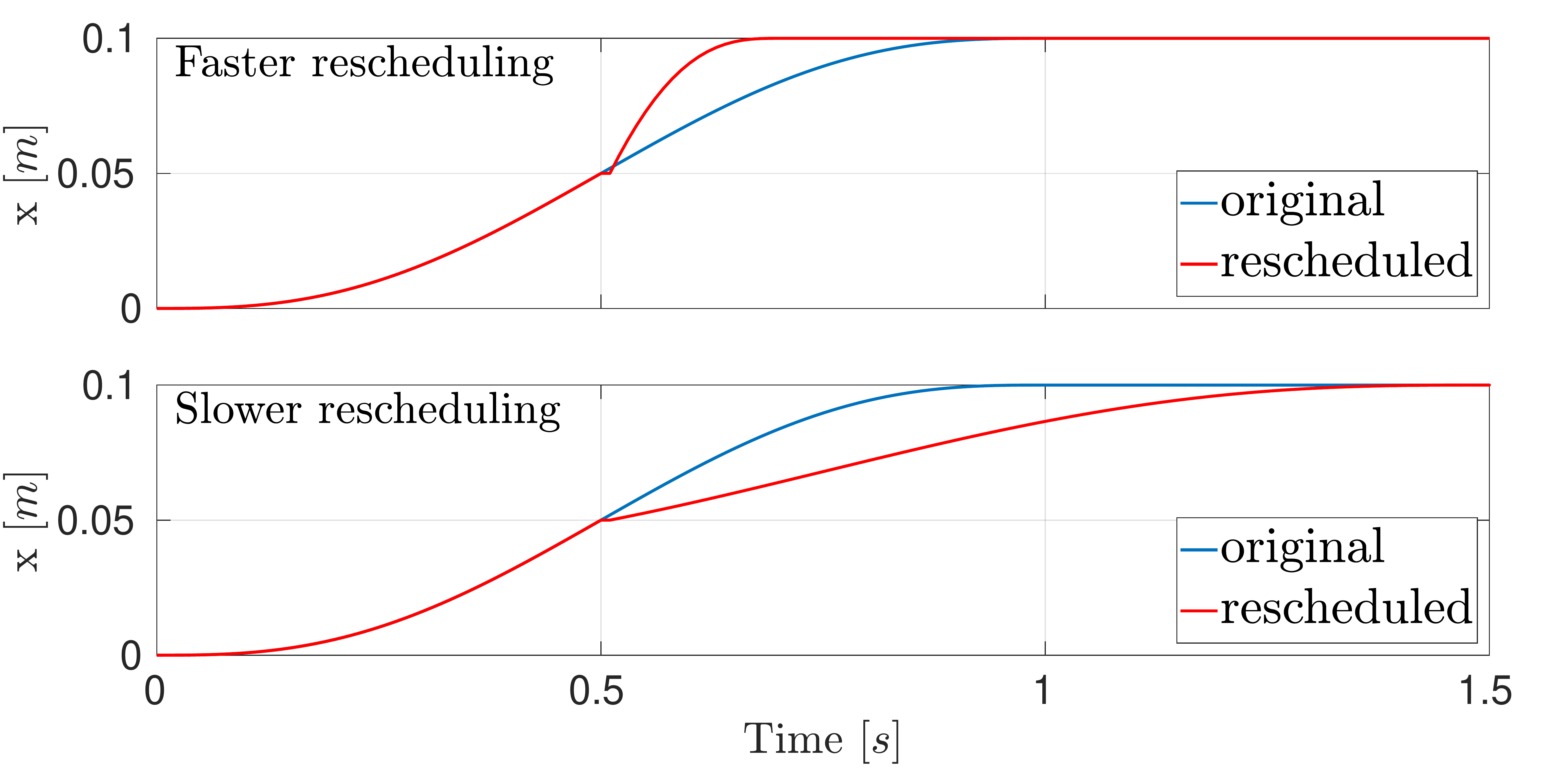}
\caption{Time rescheduling. The original trajectory (blue) at $\bar{t} = 0.5s$
is rescheduled
into a new trajectory (red) of duration of (Upper plot) $0.7s$ or (lower plot)
$1.5s$.}
\label{fig:timeRescheduling}
\end{figure}

In Fig. \ref{fig:timeRescheduling}, we show an example of time rescheduling
happening at $0.5 s$,
where the duration of an original trajectory $T_f=1 s$   is reduced to
$T_f^{'}=0.7 s$ (fast scheduling)
or increased to $T_f^{'} = 1.5s$ (slower scheduling).

\section{Reactive Behaviors}
\label{sec:reactiveFeatures}

In this section, we briefly describe the framework's reactive  modules for
robust locomotion. These modules implement strategies to mitigate the
negative effects of unpredictable events, such as: 1) slippage (Section
\ref{sec:slipDetection});
2) loss of mobility (Section \ref{sec:heightReflex});
3) frontal impacts (see Section \ref{sec:stepReflex}) and
4) unexpected contacts (\eg shin collisions, see Section
\ref{sec:shinCollision}).
\subsection{Slip Detection}
\label{sec:slipDetection}
The causes of slippage during locomotion can be divided  into three categories:
1) wrong estimation of the terrain normal; 2) wrong estimation of the friction
coefficient; 3) external disturbances.

In our previous work \cite{Focchi2018}, we addressed the first and the
second categories. In particular, we proposed  a slippage
detection algorithm which  estimates
\textit{online} the friction coefficient and the normal to the terrain. After
the estimation, the coefficient were passed to the whole-body
controller (see slip
detection and recovery module in Fig. \ref{fig:blockDiagram}).

Concerning the third category, an external push can create a loss of
contact. This can cause a sudden inwards motion of the foot. For
instance, the trunk
controller can create internal forces, depending on the regularization used.
When there is loss of contact, these internal forces can make the foot move away from the
desired position, leading to large tracking
error and catastrophic loss of balance. 
Thanks to the impedance controller (a PD running in parallel to
the Trunk Controller), the tracking error will be limited, and it will be
recovered at the next replanning stage (see
Section \ref{sec:bodyMotionPhase}).
\subsection{Height Reflex} \label{sec:heightReflex}
The \textit{height reflex} is a motion generation strategy
for all the robot's feet.
It redistributes the \textit{swing} motion onto the \textit{stance} legs, with
the final effect of ``lowering'' the  trunk
to assist the foothold \textit{searching} motion.

The \textit{height reflex} is useful when the robot is facing considerable
changes in the terrain elevation (\eg stepping down from a high platform)
\cite{focchi17iros}. In such situations, the swing leg can lose mobility,
causing issues during the subsequent steps (\eg walking with excessively
stretched legs). For this reason, the height reflex is most
likely to be  activated when stepping down.
\subsection{Step Reflex}
\label{sec:stepReflex}
The \textit{step reflex} \cite{focchi2013clawar} is a local elevation
recovery strategy, triggered in cases of frontal impacts with an obstacle
during the swing up phase.

This reflex is key in cases of visual deprivation (\eg smoky areas or thick
vegetation): it allows the robot to overcome an obstacle and establish a stable
foothold at the same time, without \textit{stumbling}.

\begin{figure}[tb]
\centering
\includegraphics[width=0.8\columnwidth]{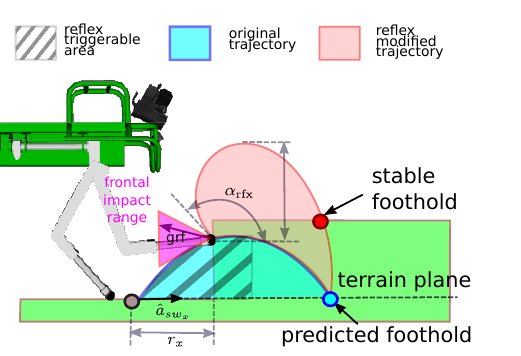}
\caption{Step Reflex. In shaded blue the original trajectory
while in shaded red is the modification due to the reflex.
The red cone, shows the range of \gls*{grf} that are eligible as frontal
impacts. Out of this cone
the \gls*{grf} will be used to trigger the stance.
Area of possible activation along the trajectory,
is depicted with stripes. }
\label{fig:stepReflex}
\end{figure}

Since the step reflex can be enabled only during the \textit{swing} phase,
it is important to set its duration $T_{\text{rfx}}$ to stop concurrently with
the end of the
default swing duration:
\begin{equation}
T_{\text{rfx}} = T_{sw} - \bar{t},
 \label{eq:step-duration}
\end{equation}
where  $\bar{t}$
is the time elapsed from the beginning of
the swing until the moment the reflex is triggered
(the searching motion will be still
possible to accommodate for further errors).

%reatraction
The angle of retraction $\alpha_{\text{rfx}}$ (see Fig.
\ref{fig:stepReflex})
depends on the distance $r_x$
(along the swing $X$ axis $\hat{a}_{{sw}_x}$ of the swing frame)
already covered by the swing foot:
\begin{equation}
r_x = \hat{a}_{{sw}_x}^T (x_{f_{sw}}^d(\batchmode{\bar{t}}) - x_{f_{sw}}^d(0)),
\label{eq:distance-covered-foot}
\end{equation}
and the reflex maximum vertical retraction $r_z$:
\begin{equation}
\label{eq:retractionAngle}
\alpha_{\text{rfx}} = \atantwo(r_z, r_x). \nonumber
\end{equation}

The maximum vertical retraction $r_z$ is an input parameter related to
the step height and the roughness of the terrain.
Note that the step reflex is also conveniently generated in the
\textit{swing frame}.

%reflex activation
The reflex can be triggered only when the impact is frontal (\gls*{grf} in the
purple cone of Fig. \ref{fig:stepReflex}), and only during
the swing \textit{up} phase (\eg before the apex point).
The reason for these constraints is twofold:
\begin{enumerate}
\item a force from a frontal impact is larger (and pointing downwards)
if the impact occurs in the swing up phase; in contrast, it is lower (and
pointing upwards) in the swing down phase. Therefore, a small and upward
force would cause an increment of false positives for a regular touchdown
detection.
\item the time available for a reflex after the swing apex is more limited and
would require significantly higher accelerations to be executed.
\end{enumerate}

\noindent \textbf{\textit{Missed reflex:}}
If a frontal impact is detected in the swing down phase,
the reflex is not triggered immediately, but it is scheduled for the beginning
of the next swing phase.

%Since the robot accumulates errors between the
%collision
%and the new rescheduled reflex, we reschedule the gait as
%well (similarly to Section \ref{sec:timeRescheduling});
%if the missed reflex is on $RF$ leg, the gait is rescheduled so that the
%next swing is $LF$ (and then again $RF$).
%As additional safety measure, we also reduce the user speed  to a safer value.

The accompanying video\footnote{Step reflex video: \url{https://www.youtube.com/watch?v=_7ud4zIt-Gw&t=2m33s}}
shows a simulation where the robot performs  a blind stair ascent, using
the heuristic-based stepping strategy as described in Section
\ref{sec:heuristicStepping}). Even if the swing leg stumbles against the step,
the robot is still able to climb the stairs, thanks to the step reflex.
In contrast, when the step reflex is \textit{disabled}, the robot
gets stuck.

\textbf{Remark:} the reflex is omni-directional and depends on the desired
locomotion speed. For instance, if the robot
walks sideways, the step will be triggered along the swing-plane, which will be
oriented laterally.

\subsection{Shin Collision}
\label{sec:shinCollision}
The foot might not be the only  point of contact with the terrain.
For certain configurations, the shin of a quadruped robot can collide
as well (as shown in this
simulation \footnote{Shin coll.
video: \url{https://www.youtube.com/watch?v=_7ud4zIt-Gw&t=4m04s}}).
Since contact forces directly influence the
dynamics of the robot,  we take this into account in the trunk
controller, \ie with an appropriate
weight redistribution after updating the number and location of the contact
points.
The detection of the contact point location can be done
either with a dedicated sensor or an estimation algorithm. On
the same line, the contact points are updated during the generation of body
trajectories.

\subsection{Experimental Results}
In this section, we present the experiments carried out on our robotic platform
\gls*{hyq}, to show the effectiveness of the proposed reactive behaviors framework in addressing
rough terrain locomotion. All the experiments have been carried out on
\gls*{hyq}, a \SI{85}{\kilo\gram}, fully-torque controlled, hydraulically
actuated quadruped robot \cite{semini11hyqdesignjsce}.

\gls*{hyq} is equipped
with a variety of sensors\footnote{For a complete description of the sensor
setup, see \cite{camurri17thesis}, Chapter 3}, including: precision joint
encoders, force/torque sensors, a depth camera (Asus Xtion), a combined (stereo
and LiDAR) vision sensor (MultiSense SL), and a tactical grade Inertial
Measurement Unit (KVH 1775).

The size of \gls*{hyq} is \SI{1.0 x 0.5 x
0.98}{\meter} (L $\times$ W $\times$ H). The leg's length ranges from
\SI{0.339}{\meter} to \SI{0.789}{\meter} and the hip-to-hip distance is
\SI{0.75}{\meter}.

\gls*{hyq} has two onboard computers: a real-time compliant
(Xenomai-Linux) used for locomotion and a non-RT machine to process vision data.
The RT PC processes the low-level controller (hydraulic
actuator controller) at \SI{1}{\kilo\hertz} and communicates with
sensors and actuators through EtherCAT. Additionally, this PC runs the
high-level controller at \SI{250}{\hertz} and the state estimation at
\SI{500}{\hertz}. The non-RT PC processes the exteroceptive sensors to generate
a 2.5-D terrain map \cite{Fankhauser2016GridMapLibrary} with
\SI{4}{\centi\meter} resolution and \SI{3 x 3}{\meter} size, surrounding the
robot.

The template terrain used for the experiments
is shown in Fig. \ref{fig:roughTerrain}. It is composed by:
1) ascending ramp; 2) a step-wise elevation change; 3) area with big stones
(diameter up to \SI{12}{\centi\meter});
4) another step-wise elevation change; 5) a descending ramp with random bricks.
Walking on stones and bricks is challenging from the locomotion point of view:
the stones can collapse or roll away, causing a loss of balance, if specific
strategies are not implemented.
%
%little description of the Experimental platform and camera
This video\footnote{Rough terrain experiments:\\
\url{https://www.youtube.com/watch?v=_7ud4zIt-Gw&t=0m10s}}
shows the robot successfully traversing the template terrain.
%demonstration of terrain adaptation
The terrain adaptation capabilities (\eg haptic feedback, searching motion)
are mostly demonstrated when the robot is walking on rocks. The
changes in elevation  (where the \textit{height reflex} is triggered) prevents
the swing leg from losing mobility.

The importance of replanning is particularly evident during the stair
descent, where the robot steps on bricks rolling away under a foothold. The
pitch error
caused by the rolling bricks is recovered in the next step. Since the
crawl is statically
stable, the robot can move in extremely cluttered environments, almost at ground
level.
In the last scene of the video, we show a simulation with the second generation of
our robots, HyQ2Max \cite{tmech16semini}. Thanks to its increased range of motion with respect to HyQ, this robot is able to crawl with a very low desired body height and is thus able to walk through \SI{45}{\centi\meter} wide duct.
%\MarcoC{Also ANYmal moves inside the duct, but crawling on the belly}

\section{Terrain Estimation}
\label{sec:terrainEstimation}
The terrain estimation module (see Fig. \ref{fig:blockDiagram})
estimates the \textit{terrain plane}, which is a linear approximation of the
terrain enclosed by the stance feet.

The inclination of the terrain plane ($\phi_t$,$\theta_t$)
is updated at each \textit{touchdown} event
(\eg a when a foot is ``sampling'' the terrain).
Typically, this is done by fitting a plane through the stance feet and
 computing the principal directions
of the matrix containing the feet positions.
The fitting plane can be found in two ways:
\begin{itemize}
\item \textbf{Vertical Fit:}  it minimizes
the ``distance'' along the \textit{vertical}  component, between the fitting
plane
($\Pi: ax+by+cz + d = 0$) and the set of points represented by the feet
position;
\item \textbf{Affine Fit:} it minimizes the Euclidean distance (along the
terrain plane normal)
between the feet and the plane.
\end{itemize}
In the first case, we have to solve a linear system,
while the second is an \textit{eigenvalue} problem.
\subsection{Vertical Fit}
\label{sec:vertical_fit}
The ``vertical'' distance can be minimized by setting the $c$
coefficient to be equal to 1 and solve a \textit{linear system}
for the $x =\mat{a & b & d}$ parameters:

\begin{equation}
x = A^{\text{\#}}b,
\label{eq:verticalFit}
\end{equation}

\noindent where  the positions of the stance feet have been collected in:
\begin{equation}
A = \mat{ x_{f_{1x}}  & x_{f_{1y}} & 1 \\
					  x_{f_{2x}}  & x_{f_{2y}} & 1 \\
					  x_{f_{3x}}  & x_{f_{3y}} & 1 \\
					  x_{f_{4x}}  & x_{f_{4y}} & 1 },
\quad
b = \mat{ -x_{f_{1z}}  \\
					  -x_{f_{2z}}   \\
					  -x_{f_{3z}}   \\
					  -x_{f_{4z}}  }.
\label{eq:verticalFitMatrix}
\end{equation}
and where $[\cdot]^{\#}$ is the Moore-Penrose pseudoinverse operator.
Then, the normal  to the terrain plane $n_t$ and
the correspondding roll/pitch angles $\phi_t, \theta_t$ (in ZYX convention),
can be obtained as\footnote{Note that the normal
vector [a,b,1] should be normalized  for the computation of $\phi_t,
\theta_t$.}:
\begin{align}
\label{eq:terrainParameters}
n_t &= \mat{a &b& 1}^T/ \Vert \mat{a&b&1}\Vert, \nonumber \\
\theta_t &= \atan(n_{t_x}/n_{t_z}),\\
\phi_t &= \atan(-n_{t_y}\sin(\theta_t )/n_{t_x}). \nonumber
\end{align}

\subsection{Affine Fit}
\label{sec:affine_fit}
In the affine case, we first need to reduce the
feet samples by subtracting
their average $\bar{x}$ (belonging to the fitting plane):
\begin{equation}
R = \mat{ x_{f_{1}}^T  - \bar{x}^T \\
		 x_{f_{2}}^T - \bar{x}^T \\
		  x_{f_{3}}^T - \bar{x}^T \\
		  x_{f_{4}}^T  - \bar{x}^T},
\quad  \bar{x} =
\frac{1}{4}\sum_{i=1}^4 x_{f_{i}}.
\label{eq:affineFit}
\end{equation}
The principal directions of this set of samples
are the eigenvectors of $R^TR$:

\begin{align}
\mat{V & D}& = eig\left(R^TR \right)  \nonumber \\
n_t &= S_1V
\label{eq:affineFit2}
\end{align}

\noindent where $V$ is the matrix of the eigenvectors and $D$ the diagonal
matrix of the eigenvalues of $R^TR$.

We can obtain the terrain normal $n_t$ by extracting the first column from  the
eigenvector matrix
(\eg the eigenvector associated to the smallest eigenvalue).
Then, similarly to the vertical fit case, \eref{eq:terrainParameters}
can be applied to find the terrain parameters. Note that the result is slightly
different from the vertical fit case. Indeed, in the affine fit case, we
minimize the Euclidean distance, while in the vertical fit case we minimize
the distance along the $z$ direction. On the other hand, the two approaches give
the same result if the feet are coplanar.

It is noteworthy that the affine approach does not provide
meaningful results when $R^TR$ is rank deficient. However, this happens only
when (at least) three feet are aligned, which is a very unlikely situation.
\subsection{Correction for Rough Terrain}
\label{sec:smartTerrainEstimation}
There are some situations where just fitting
an average plane is not the ideal thing to
do during a  statically stable motion.
For instance, when the robot has three feet on the ground
and one on a pallet, the \textit{average} (fitting) plane is not horizontal.
In this case, moving the \gls*{com} projection along the terrain plane
(to enter in the support triangle) results in an inconvenient ``up and down''
motion. This happens because the robot tries to follow the orientation of
the terrain plane with its torso.
In this case, it would be preferable to keep the posture horizontal
until \textit{at least} two feet are on the pallet.

In this section, we propose a more robust
implementation of the terrain estimation strategy, which allows to
address these particular situations.
The idea is to have the terrain plane fitting the
\textit{subset} of the stance feet that are closer to be \textit{coplanar}.
In this case, the influence of the ``outlier'' foot (\eg the one on the pallet)
would be reduced. For  more dynamic gaits, where the \gls*{com} trajectory is
not planned (\eg like trotting \cite{Barasuol2013}),
a  low-pass filtering of the terrain estimate would be sufficient to mitigate
the
effect of the outliers.
However, since the crawl makes heavily use of the \textit{terrain plane} to
change the pose of the robot, a different strategy has be adopted.

A preliminary step (after  computing the terrain normal $n_t$ as in
Section \ref{sec:vertical_fit})
is to compute the norm of the least square errors vector. This can be easily
done by
exploiting the matrix computed in \eref{eq:verticalFitMatrix}: $e_{LS}= \Vert
Ax-b \Vert_2$.
If $e_{LS}$ (Least Square error) is bigger than a certain (user defined)
threshold, it means
that the feet are not \textit{coplanar},
and $n_t$ should be corrected.

\begin{figure}[tb]
	\centering
	\includegraphics[width=0.6\columnwidth]{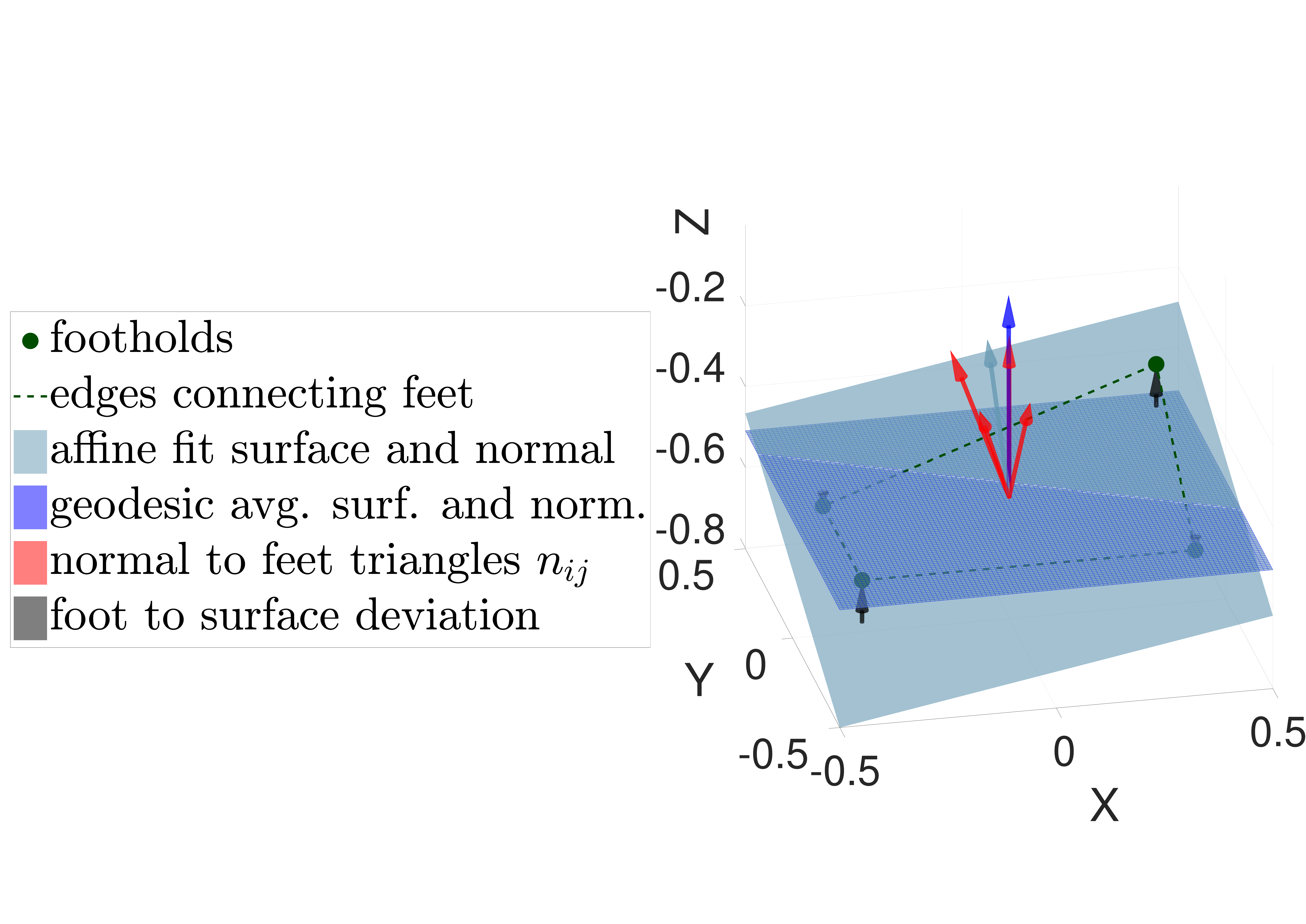}
	\includegraphics[width=0.38\columnwidth]{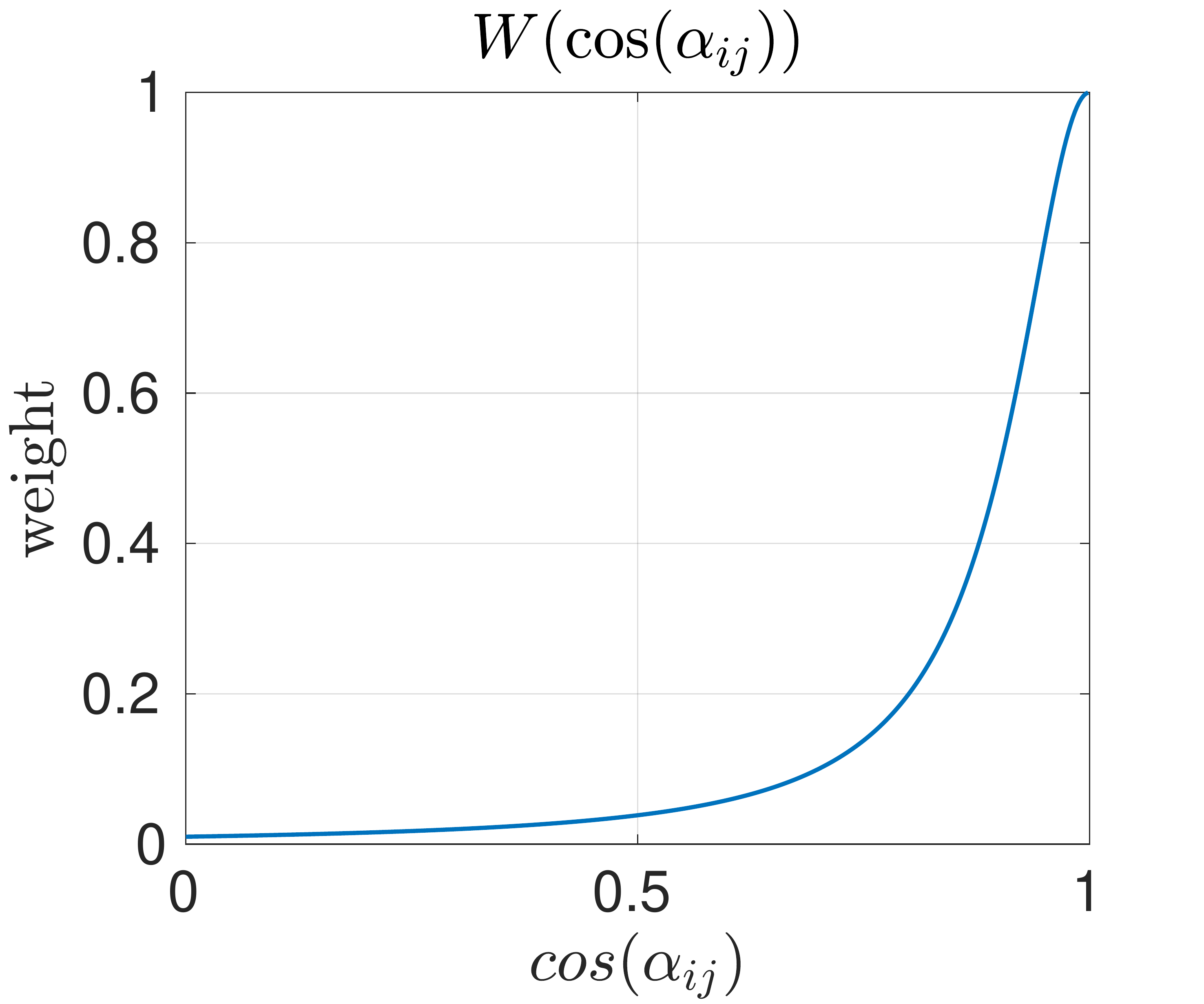}
	\caption{Smart height correction.  (Left) the blue arrow represent the
corrected normal while the grey arrow the not corrected one (computed with the
affine fit method). In this case the corrected normal is almost coincident to one of the red arrows
representing the normals $n_{ij}$ to the triangles defined by three stance feet;
(right) weight function (W)  of the angular distance (\eg cosine) w.r.t
$n_{t_{old}}$.}
	\label{fig:terrainCorrection}
\end{figure}

First, we compute the normal vectors in common to
all the combinations of two adjacent edges ($l_i$, $l_j$),
(sorted in \gls*{ccw} order, see Fig. \ref{fig:terrainCorrection}):
\begin{align}
n_{ij} = l_i\times l_j   \quad (i,j) \in C,
\label{eq:normal}
\end{align}
where $C$ is the set of all the combinations of two
adjacent edges (sorted in \gls*{ccw} order).
In our case, $C$  has 4 elements.
Since any couple of (intersecting or parallel) lines defines a plane, we
associate each normal $n_{ij}$ to one support triangle (\eg two edges of the
support polygon, see Fig. \ref{fig:terrainCorrection}~(left)).

The corrected terrain normal is
a \textit{weighted} average of $n_{ij}$, where the weights are
inversely proportional to the distance  of each normal $n_{ij}$ from
the terrain plane normal, computed at the previous
touchdown event  $n_{t_{_\textup{old}}}$.
\begin{align}
&cos(\alpha_{ij})=  n_{t_{old}}^Tn_{ij}.
\label{eq:angularDistance}
\end{align}

We compute the weight $w_{ij}$ associated  to each normal $n_{ij}$
through the nonlinear function $W(\cdot) \in \Rnum \rightarrow \Rnum$ (Fig.
\ref{fig:terrainCorrection} (right)),
in accordance to its angular distance from  $n_{t_{_\textup{old}}}$.

\begin{align}
&W(x)  = 1/\left(1 + s (x - 1)^p\right),  \nonumber  \\
&w_k = W(cos(\alpha_{ij})),
\label{eq:weight}
\end{align}
where $k$ is the index of the $ij-th$ element of the set $C$;
and $s$ is a sensitivity factor proportional to the LS error $e_{LS}$.

According to \eref{eq:weight}, the normals closer
to $n_{t{_\textup{old}}}$ (\eg cosine close to 1) are assigned a bigger weight
(the weight is bounded to 1 by construction of $W(\cdot)$).
The exponent $p$ allows us to adjust
the degree of nonlinearity of $W(\cdot)$
and the degree of correction (in our case $p=2$ was sufficient).

Since the normals $n_{ij}$ are
directly affected by the position of their
corresponding feet on the pallet, a weighted average  of the normals (with
weights
inversely proportional to the distance from the previous
 estimation)
allows to naturally reduce the influence
of the ``outlier'' foot.
This discourages the terrain estimator from modifying too much the previous
estimate when a foot position is far away from the previous fitting plane.

Note that, since we are aiming at  ``averaging''
orientations, we need to perform a \gls*{slerp}
using geodesic curves \cite{Gramkow2001} (see $Appendix$ $A$).

\subsection{Experimental Results}
The terrain estimation
video\footnote{Terrain estimation video: \url{https://www.youtube.com/watch?v=_7ud4zIt-Gw&t=5m01s}}
shows how the smart terrain estimator
improves locomotion by removing undesired up/down motions.

With reference to our initial example, as long as the robot
has one foot on the pallet,
the normal is closer to the one provided by the other
stance feet --- which in turn are closer to the
previous estimation with all the
feet   on the flat ground (see Fig. \ref{fig:terrainCorrection}).

When the robot steps with the two lateral feet on the pallet,
the fitting error $e_{LS}$ is reduced (the feet are more
coplanar) and an inclined terrain plane is estimated.

The approach can address situations
with non coplanar feet (\eg support has a ``diamond'' shape, shown later in the
video).
In this case, the terms from the feet on the pallet cancel each
other,
keeping the previous terrain plane estimate unchanged,
which is the desirable behavior. 
For the single pallet example,  we have $e_{LS} =   0.02$
(for the ``diamond'' shape $e_{LS} = 0.031$),
therefore we set the threshold of
intervention for the terrain correction to $0.002$.

%sensitivity
Since we want a stronger correction when the $LS$ error increases, we
make the sensitivity factor $s$ of the function $W(\cdot)$ proportional to
$e_{LS}$.

\section{Stair climbing}\label{sec:stairClimbing}
Stair climbing is an essential skill in the
for a legged robot.
Indeed, a \textit{versatile} legged robot  should be able to
address both \textit{unstructured} and \textit{structured} environments,
such as the one we can find in a disaster scenario.

The problem of stair climbing can be addressed through foothold planning,  to
avoid collisions with the step edges.
An optimization taking into account the full kinematic of the robot could
possibly solve the problem, but it is  currently hard to  be performed
\textit{online}.

Moreover, as previously mentioned, a full optimization approach
that plans for a whole staircase is prone to tracking errors, which
might eventually end up in missing a step and fall.

An alternative strategy is to conservatively select
the foothold in the middle of the step (depth-wise). However,
depending on the inclination of
the stairs and on the step-size, the robot can end up in
inconvenient configurations from
the kinematic point of view (\eg with degeneration of
the support triangle and associated loss of mobility).

In the case of \gls*{hyq}, the kinematic limits at
the Hip-Flexion-Extension joints (\ie hip joints rotating around the
$Y$-axis, see \cite{semini11hyqdesignjsce}) are likely to be hit
during the  \textit{body motion} phase\footnote{We adopt a
``telescopic strut'' strategy as in \cite{Gehring2016},
which means that, on an inclined terrain, the vector between each stance foot
and the corresponding hip aiming to be maintained parallel to gravity. 
On one hand, this
improves the margin for a static equilibrium.
 On the other hand, for high
stair inclinations, it can result in bigger joint motions,
where the kinematic limits are most likely hit.}. 

According to our experience,
keeping the joint posture as close as possible to the \textit{default}
configuration\footnote{As the one shown in Fig.
\ref{fig:roughTerrain} (left),
where the joints are in the middle of their range of motion} it
improves mobility, and is an important factor
for the robustness of locomotion.
Our heuristic approach, rather than avoiding
potentially dangerous situations (\eg missed steps, collisions),
aims to be robust enough to cope with them.
%
%Our locomotion approach, since it plans reactively
%the (default) steps about the hips
%follows this line of thinking trying to maintain
%the robot in a convenient kinematic configuration.
%\MarcoC{The sentence above is very difficult to understand. I would REMOVE it.}

In particular, we show that our
vision-based stepping approach, with
minor modifications (see section \ref{sec:stairLocomotionMode}),
is sufficient to successfully climb up/down industrial-size stairs. In case the
\textit{vision based swing strategy} is not sufficient to avoid frontal impacts
(\eg against one step), the \textit{step reflex} is triggered to achieve a
stable foothold and prevent the robot from getting stuck.
\subsection{Influence of the knee configuration}
% knee configuration influence
The probability of ``getting stuck'' when climbing stairs depends
on the knee configuration (bent backwards or forward).
In particular, when the leg has a \gls*{KBB}
configuration,
it is more prone to
have shin collision when climbing \textit{downstairs}.
%knee bent forward
Conversely, a \gls*{KBF} configuration increases
the risk of shin collision when
climbing \textit{upstairs}. These are important aspects for the design of
robots that are expected to climb stairs.
\begin{figure}[tb]
	\centering
	\includegraphics[width=0.48\columnwidth]{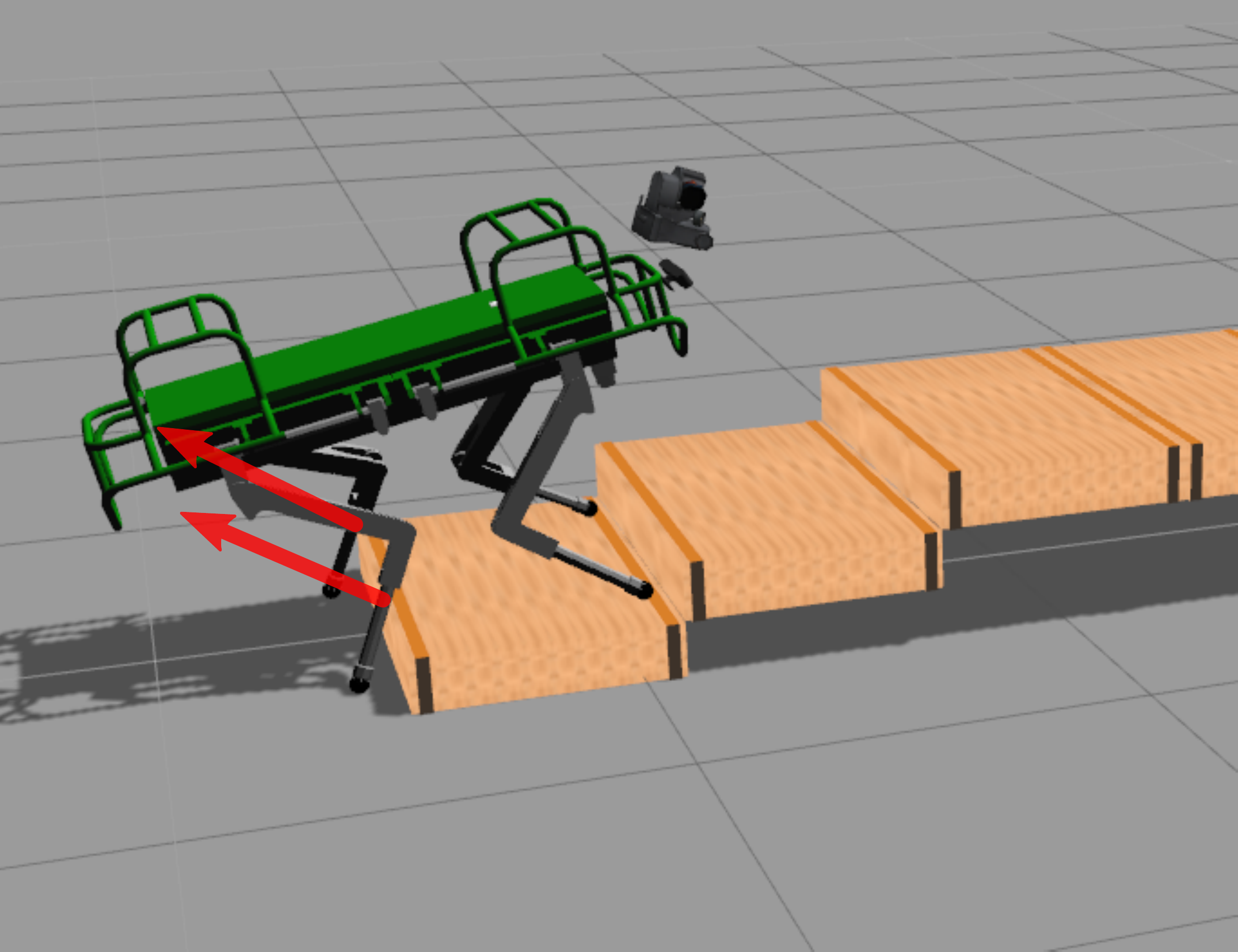}
	\includegraphics[width=0.5\columnwidth]{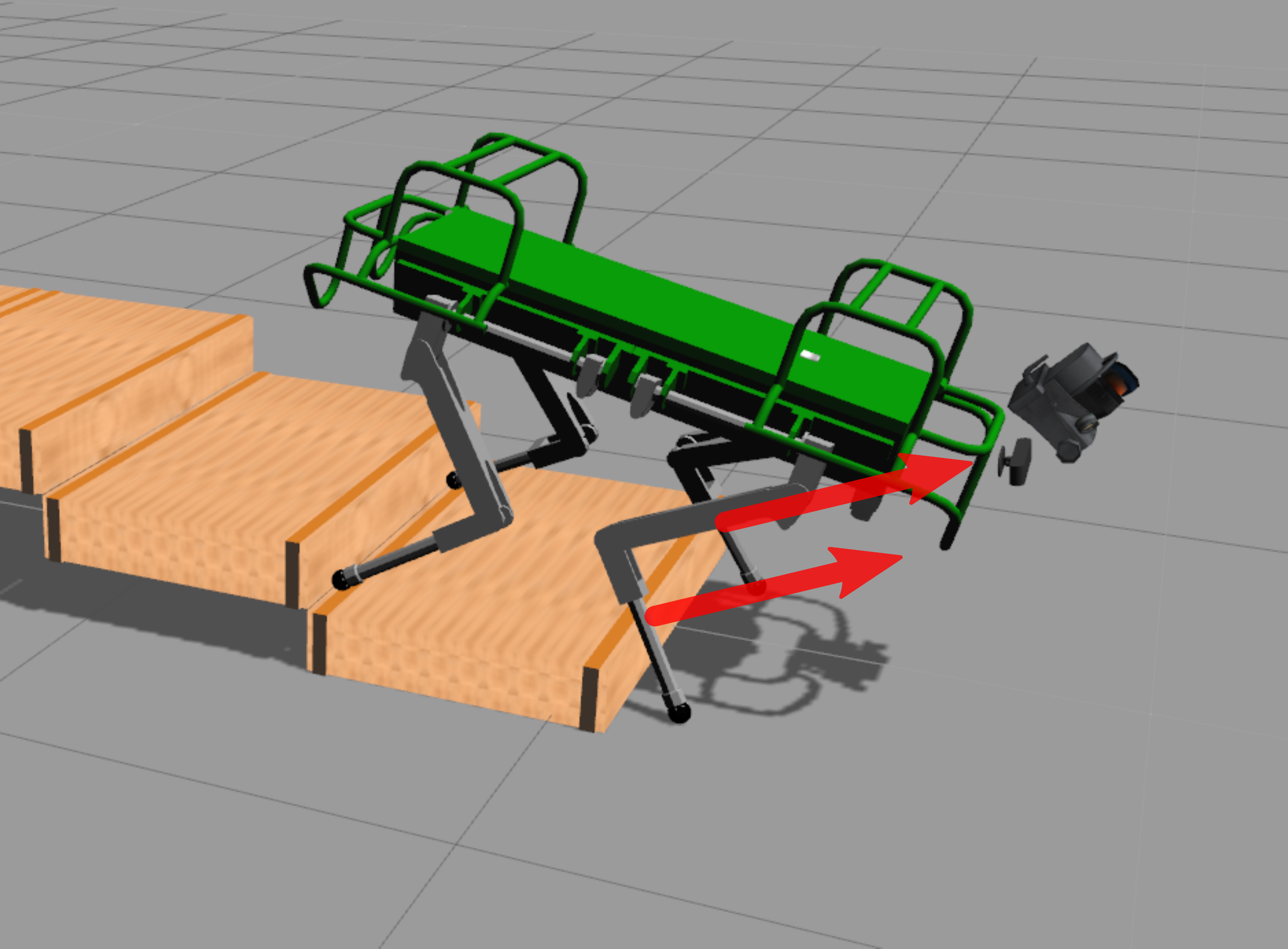}
	\caption{Comparison of \gls*{KBB} and \gls*{KBF} configurations
	during stair climbing. In the \gls*{KBF} while climbing up the
\gls*{grf}
	point against the direction of motion making the robot get stuck.}
	\label{fig:kneeConfigs}
\end{figure}

Note that the contact forces at the
shin are \textit{in} the direction of motion when
the configuration is \gls*{KBB} and the robot is climbing downstairs,
whereas they point \textit{against} the motion with the \gls*{KBF}
configuration and the robot climbs upstairs (see Fig. \ref{fig:kneeConfigs}).
In the former case we might have slippage, but the robot eventually moves
forward,
while in the latter case the robot might get stuck or fall backwards, unless
these situation are properly dealt with (see Section
\ref{sec:shinCollision}).

\subsection{Stair locomotion mode}
\label{sec:stairLocomotionMode}
The \textit{stair locomotion mode} can be triggered either by the user or by
a dedicated stair detection algorithm \cite{Oswald2011}.
It consists in the activation
of three  features (relevant for the task of climbing stairs) on top
of the vision-based stepping strategy:
\begin{enumerate}
\item \textbf{Rescheduling of the gait sequence:}
the  kinematic configurations that cause  mobility loss are undesired. To
avoid them, it is important to climb stairs
with both front feet (or back feet) on the same step.
Specifically, the \textit{next} foot target
is checked at each touchdown (\ie before the \textit{move body} phase).
If this is on a different height than the foot on the opposite side,
we perform a rescheduling of the gait sequence.

For instance, if the last swing foot was
the right front ($RF$) then, according to our \textit{default} configuration
($RH, RF, LH, LF$)\footnote{This sequence is the one animals employ that reduces
the backward motions \cite{2007_pongas}.}
the next leg to swing would be the $LH$. However, if the left-front ($LF$) foot
is on a different height (\eg still on the previous step),
the whole sequence is rescheduled to move the $LF$ instead.
The same applies for the back feet.

\item \textbf{Conservative stepping:} taking inspiration from the ideas
presented in \cite{barasuol15iros},
this module corrects the foot location to step far away from an edge.
The terrain flatness is checked along the direction of
motion and the foothold is corrected (along the swing plane)
in order to  place it in a more conservative location (\ie away from the step
edge).

\item \textbf{Clearance optimization}:
this feature (presented in Section \ref{sec:clearanceOptimization}) is
useful to avoid the chance of stumbling against the step's edge.
It adjusts the swing apex and the step height to maximize the clearance
from the step.
\end{enumerate}
\subsection{Experimental Results}
We successfully applied the reactive modules of our framework to the problem
of climbing up (and down) stairs, in simulation, with
our quadruped robot,
HyQ\footnote{Stair climb video:\\
\url{https://www.youtube.com/watch?v=_7ud4zIt-Gw&t=7m13s}}.
In the video, we show that HyQ is able to climb up and down industrial size
stairs (step raise \SI{14}{\centi\meter})
and climbing up a staircase with a \SI{90}{\degree} turn.

The approach is generic enough to be used also with
irregular stair patterns (different step raise)  and turning stairs.
The user provides only a reference speed and heading.

In the  simulation video we show the advantage of activating the stair
locomotion mode
and the importance of using a vision based stepping strategy.

In the \SI{90}{\degree} staircase, we demonstrate omni-directional
capability of our statically stable approach, which allows to move backwards on
the staircase.

\section{Momentum Based Disturbance Observer}
\label{sec:momentumBasedObserver}
%motivation
A significant source of error
might come from unmodeled disturbances, such as an external push.
Specifically, in the case
of a model-based controller
(\eg our Whole Body controller \cite{focchi2017auro}),
inaccurate model parameters cause a wrong prediction of the
joint torques. This shifts the responsibility of the control  to the
feedback-based
controllers, thus increasing tracking errors and delays.
%why online id
However, if a proper identification is carried out \textit{offline},
these model inaccuracies are mainly restricted to the trunk.
Indeed, in the case of our quadruped robot,
the leg inertia does not change significantly,
but the trunk parameters are instead strongly dependent  on
robot payload (\eg a backpack, an additional computer,
different sets of cameras for perception, \emph{etc.}).
%online id

In \cite{tournois17iros},  we presented
a \textit{recursive} strategy which
performs \textit{online} payload identification to estimate the new
\gls*{com} position of the robot's trunk.
The updated model is then used
for a more accurate inversion of the dynamics \cite{focchi2017auro,
mastalli17icra}.
%
%motivation for momentum based disturbance observers
Even though this approach is effective to detect
constant payload changes,
it is not convenient to
estimate \textit{time-varying}
unknown external forces, which might change
both in \textit{direction}  and  \textit{intensity}.
Indeed, this kind of disturbances can dramatically
increase the tracking errors and jeopardize the
locomotion, unless they are compensated \textit{online}.

\begin{figure}[tb]
	\centering
\includegraphics[width=1.0\columnwidth]{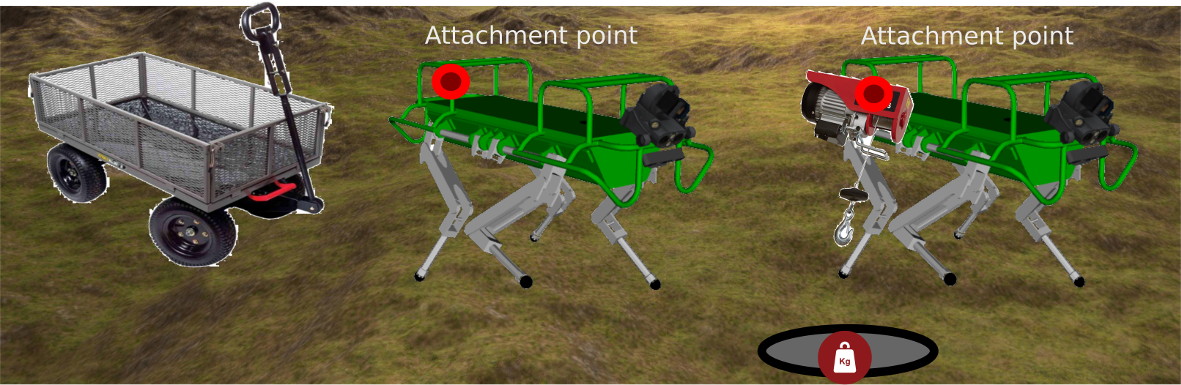}
	\caption{Use cases for load estimation: (left) the robot is pulling a
cart,		(right) the robot is pulling up a payload.}
	\label{fig:wheelBarrowTemplates}
\end{figure}

In the following, we mention two scenarios  where an
external disturbance observer is useful:
1) the robot is required to pull a cart or a load
(\eg with some delicate material inside); and 2) the robot is requested to pull
up
a payload from underneath with a hoist mounted on the
torso (see \ref{fig:wheelBarrowTemplates}).

In this section, we present a \gls*{mbdo} able to estimate an external
\textit{wrench}
(\eg cumulative effect of all the disturbance forces and moments).
We also show how this \textit{wrench}  is compensated during the locomotion,
thus improving tracking accuracy and locomotion stability.

Our approach builds on top of the one described in \cite{englsberger17iros},
which is designed to estimate an external \textit{linear} force acting on the
robot base. We extended this work to the angular case,
estimating the full wrench at the \gls*{com}.

Since our robot has a non negligible angular dynamics, estimating only a
\textit{linear} force has limited effectiveness.
Unless the force is applied at the \gls*{com},
the \textit{application point}, then  the moment about the
\gls*{com} must be  worked out.
Our approach is more general as it does not require the application point to
estimate the full wrench.

%idea for estimation
The idea underlying  the estimation is that any external
\textit{wrench}\footnote{Henceforth, for simplicity,
we talk about coordinate vectors and not spatial vectors,
that is why $W_{ext} \in \Rnum^6$ 	rather than $W_{ext} \in F^6$
\cite{Featherstone2008}} $W_{ext} \in \Rnum^6$
has
an influence on the \textit{centroidal}
momentum (\ie linear and angular momentum at the \gls*{com}) \cite{Orin2013}.

We observe that the discrepancy between the
predicted (centroidal) spatial momentum $\hat{h}(t) \in \Rnum{^6}$,
 based on \textit{known} forces (\eg gravity and \gls*{grf}), and
the measured one $h(t)$, based on proprioceptive
estimation of the \gls*{com} twist\footnote{The twist (\ie 6D spatial
velocity) is
usually computed by a state estimator, which merges \gls*{IMU},
encoder and force/torque sensors \cite{camurri17ral}.},
in absence of modeling
errors is caused only by an external wrench $W_{ext}$\footnote{It is well known that  it is impossible to distinguish
	between a \gls*{com} offset and an external wrench
\cite{rotella18thesis}.
	Therefore our starting assumption is that 
	there are no modeling errors (\ie a preliminary trunk \gls*{com}
	identification has been carried out previously
	using \cite{tournois17iros}).}.
We can exploit this fact to design an observer.

To obtain a prediction of $\hat{h}(t) = \mat{\hat{p}_G(t) & \hat{k}_G(t)}$,
we exploit  the centroidal dynamics (\eg Newton-Euler
equations)\cite{Orin2013}:
\begin{equation}
\left\{
\begin{array}{ll}
\hat{\dot{p}}_G(t) = 	\underbrace{mg + \sum_{i=1}^c f_i(t)}_{f_{\text{known}}}
  +
\hat{f}_{ext}(t) \\
\hat{\dot{k}}_G(t)  =  \underbrace{\sum_{i=1}^c (x_{f_i}(t) - x_c(t)) \times
f_i(t)}_{\tau_{\text{known}}}   +  \hat{\tau}_{ext}(t) 
% I_{com} \dot{\omega} + \omega \times I_{com} \omega
\end{array}
\right.
\label{eq:centroidalDyn}
\end{equation}
where   $\hat{\dot{p}}_G(t) \in \Rnum^3$ and $\hat{\dot{k}}_G(t) \in
\Rnum^3$
are the linear and angular momentum rate, respectively;
$\hat{W}_{ext}(t) =\mat{\hat{f}_{ext}(t) &\hat{\tau}_{ext}(t)}$ is a prediction
of
the
wrench disturbance at time $t$, expressed at the \gls*{com} point.

%\footnote{Note
%that the disturbance
%wrench is a pure force with an application point $x_a$,  thus
%$\hat{\tau}_{ext}
%= x_a \times \hat{f}_{ext}$, while $f_{ext}$ does not change because it is a
% free vector.}.

Starting from an initial measure of the momentum $h_0= \mat{p_{G0}& k_{G0}} =
\mat{ m\dot{x}_{com}(0) &I_{com}(0)\omega(0)}$,
we can get the predicted $\hat{h}(t)$, at a given time $t$, by integration
of \eref{eq:centroidalDyn}:

\begin{equation}
\left\{
\begin{array}{ll}
\hat{p}_G(t)&= p_0 + \int^t_0 \left( mg + \sum_{i=1}^{c_{st}} f_i(t)   +
\hat{f}_{ext}(t) \right)\,dt \\
\hat{k}_G(t)&= k_0 + \int^t_0 (  \sum_{i=1}^{c_{st}} (x_{f_i}(t) - x_{com}(t))
\times f_i(t)   +\\ &\quad \quad \quad \quad \quad \quad \quad \quad\quad \quad
\quad \quad\quad \quad\quad\hat{\tau}_{ext}(t) )\,dt
\end{array}
\right.
\label{eq:momentumObservation}
\end{equation}

Then, the discrepancy between  the measured and the predicted momentum
can be used to  estimate the external wrench disturbance, leading to the
following observer set of equations:
\begin{equation}
\left\{
\begin{array}{ll}
\hat{f}_{ext}(t) &= G_{\text{lin}}\left( m \dot{x}_{com}(t) -  \hat{p_G}(t)
\right)
\\
\hat{\tau}_{ext}(t) &= G_{\text{ang}}\left( I_{com}(t) \omega(t)  -
\hat{k_G}(t)
\right )
\end{array}
\right.
\label{eq:updateEquation}
\end{equation}
\noindent where the gains $G_{\text{lin}}, G_{\text{ang}} \in \Rnum^{3\times3}$
are
user-defined positive definite  matrices,
which describe the observer's dynamics, and $I_{com}(t) \in \Rnum^{3 \times 3}$
is the rotational inertia
of the robot (as a rigid body) computed at time $t$.
On the other hand, \eref{eq:centroidalDyn} can be rewritten using spatial
algebra \cite{Featherstone2008},
considering the whole robot as a rigid body:
\begin{equation}
\dot{h} = \frac{d}{dt}(\bar{I}_{com}v) = \bar{I}_{com}\dot{v} + v\times^{\*}
\bar{I}_{com}v
= W_{\text{known}} + \hat{W}_{ext}
\label{eq:centroidalDynSpatial}
\end{equation}

\noindent where $v =  \mat{\dot{x}_{com} & \omega }\in \Rnum^6 $ is the measured
\gls*{com} twist composed
of \gls*{com} linear velocity and robot angular velocity (for simplicity of
notation, we omit henceforth the dependency on $t$);
$\bar{I}_{com}\in \Rnum^{6 \times 6}$ is the composite rigid body inertia
(expressed in an inertial frame attached to the \gls*{com}), evaluated at
each loop at the actual configuration of  the robot;
$W_{\text{known}} = \mat{ f_{\text{known}} & \tau_{\text{known} } }$
is the
wrench due to contacts and gravity.

Using \eref{eq:centroidalDynSpatial}, it is possible to formulate a wrench
observer
where the nonlinear term $v\times^{\*} \bar{I}_{com}v$ is compensated:

\begin{equation}
\left\{
\begin{array}{ll}
\bar{I} \hat{v} &= \bar{I}_0v_0 + \int^t_0 \left(W_{\text{known}} +
\hat{W}_{ext} -v\times^{\*} \bar{I}_{com}v  \right)\,dt \\
\hat{W}_{ext} &= G\bar{I}\left( v  -  \hat{v} \right )
\end{array}
\right.
\label{eq:updateEquationSpatial}
\end{equation}

\noindent where $G = \diag(G_{\text{lin}}, G_{\text{ang}})\in \Rnum^{6\times6} $
is the observer gain matrix.

At each control loop,  after the \textit{estimation} step, we perform
\textit{online}
compensation
of the estimated disturbance wrench $\hat{W}_{ext}$  in the whole-body Trunk
Controller \cite{focchi2017auro} (see Fig. \ref{fig:blockDiagram}):
\begin{equation}
W^d = W_{vm} + W_g - \hat{W}_{ext}
\label{eq:desWrench}
\end{equation}
\noindent where $W^d \in \Rnum^6$ is the desired wrench (expressed at the
\gls*{com}), mapped to desired joint torques
by the Trunk Controller; $W_g, W_{vm}  \in \Rnum^6$ are the
gravity compensation wrench and the virtual model
attractor (which tracks a desired \gls*{com} trajectory) wrench,
respectively.

\subsection{ZMP compensation}
Compensating for the external wrench is not sufficient to achieve stable
locomotion.
During a static crawl, the accelerations are typically small, and the
\gls*{zmp} mostly coincides with the projection of the \gls*{com}
on the support polygon. However, this does not hold if an external disturbance
is present. In this case, the \gls*{zmp} can be shifted. If this is not
properly accounted for in the body motion planning, the locomotion
stability might be at risk.

\begin{figure}[tb]
	\centering
	\includegraphics[width=1.0\columnwidth]{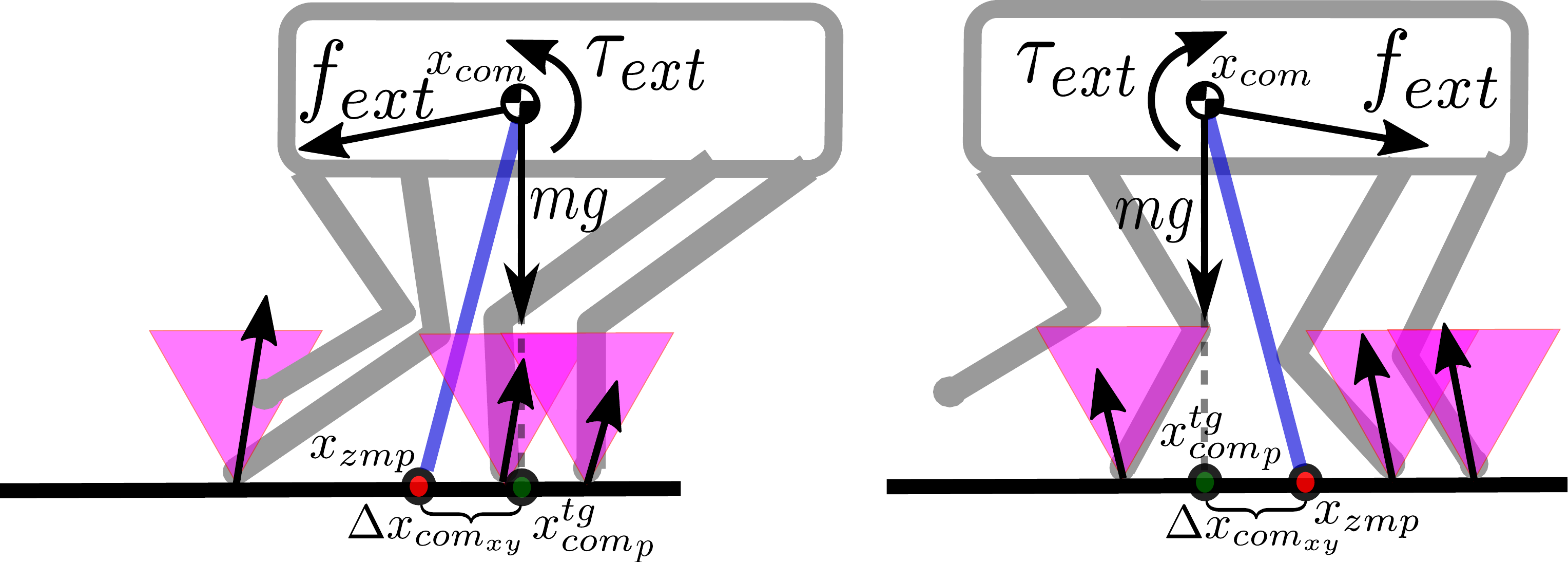}
	\caption{Diagram for the computation of the \gls*{zmp} due to external
disturbances.
	Left and right figures show different values of $\Delta x_{{com}_{xy}}$
	for different external disturbances.}
	\label{fig:zmpOffset}
\end{figure}

Knowing  that the
\gls*{zmp}
is the point on the support polygon (or, better, the line) where the tangential
moments nullify, the shift $\Delta x_{com}$ can be estimated by computing the
equilibrium of moments about this point (see Fig. \ref{fig:zmpOffset}):
\begin{equation}
\underbrace{(x_{com} - x_{zmp})}_{\Delta x_{com}} \times (mg + f_{ext}) +
\tau_{ext} = 0
\label{eq:zmpOffset}
\end{equation}
where $\Delta x_{com}$ is the vector going from  the \gls*{zmp} to the
\gls*{com}\footnote{Note that only gravity and the external wrench have an
influence on
$\Delta x_{{com}_{x,y}}$, because the
resultant of the \gls*{grf} passes through the \gls*{zmp}
point, by definition.}.
Rewriting \eref{eq:zmpOffset} in an explicit form\footnote{Note that computing
this equations as $\Delta x_{com}= [mg + f_{ext}]_{\times}\tau_{ext}$ where
$[\cdot]_{\times}
$ is the skew symmetric operator associated to the cross product, returns
inaccurate results because   $[\cdot]_{\times}$
is rank deficient.}, we get \cite{popovic2005ground}:

\begin{equation}
\left\{
\begin{array}{ll}
\Delta x_{{com}_x} = \frac{1}{f_{{ext}_z} - mg} \left[ f_{{ext}_x} (x_{{com}_z}
-
x_{{zmp}_z})  + \tau_{{ext}_y} \right]\\
\Delta x_{{com}_y} = \frac{1}{f_{{ext}_z} - mg} \left[ f_{{ext}_y} (x_{{com}_z}
-
x_{{zmp}_z})  - \tau_{{ext}_x} \right]
\end{array}
\right.
\label{eq:zmpOffsetExplicit}
\end{equation}

\noindent then, the new target computed at the beginning of the base motion
will account for this term:
\begin{equation}
x_{{com}_{x,y}}^{tg} = x_{{com}_{x,y}}^{tg} + \Delta x_{{com}_{x,y}}
\label{eq:targetBary}
\end{equation}
%
%application point estimation TODO
%
\subsection{Stability Issues}
Any observer/state feedback arrangement can lead to some stability issues if
the gains are not set properly
(\eg by separation principle). We did not  carried out a
system stability analysis, since a proper evaluation of the stability region
(and an improved
implementation taking care of this aspects) is an ongoing work
and it is out of the scope of this paper.
However, we noticed that there are some combinations of gains for which
the system becomes unstable.

In the experiments, we decided to be conservative and set lower gains  than in
simulation. We also did not see a significant  improvement in using  the
implementation \eref{eq:updateEquationSpatial} instead of
\eref{eq:updateEquation}.

\subsection{Simulations}
To evaluate the quality of the \textit{estimation}, we inject in simulation a
known disturbance: a pure force to the back of the robot (with application
point ${}_bx_p  = \mat{-0.6, 0.0, 0.08} m$, expressed in the base frame),
while the robot is standing still.
Specifically, we generate a time-varying perturbation force $f_{ext}$
with random stepwise changes both in  \textit{magnitude} (between
\SI{40}{\newton} and \SI{200}{\newton}) and \textit{direction},
with additive white Gaussian noise $n \in \mathcal{N}(0, 20) \,\si{\newton}$.
The  gains used in the observer are the following: $G_{ang} =
\diag(100,100,100)$, $G_{ang} = \diag(10,10,10)$.

\begin{figure}[tb]
	\centering
	\includegraphics[width=0.8\columnwidth]{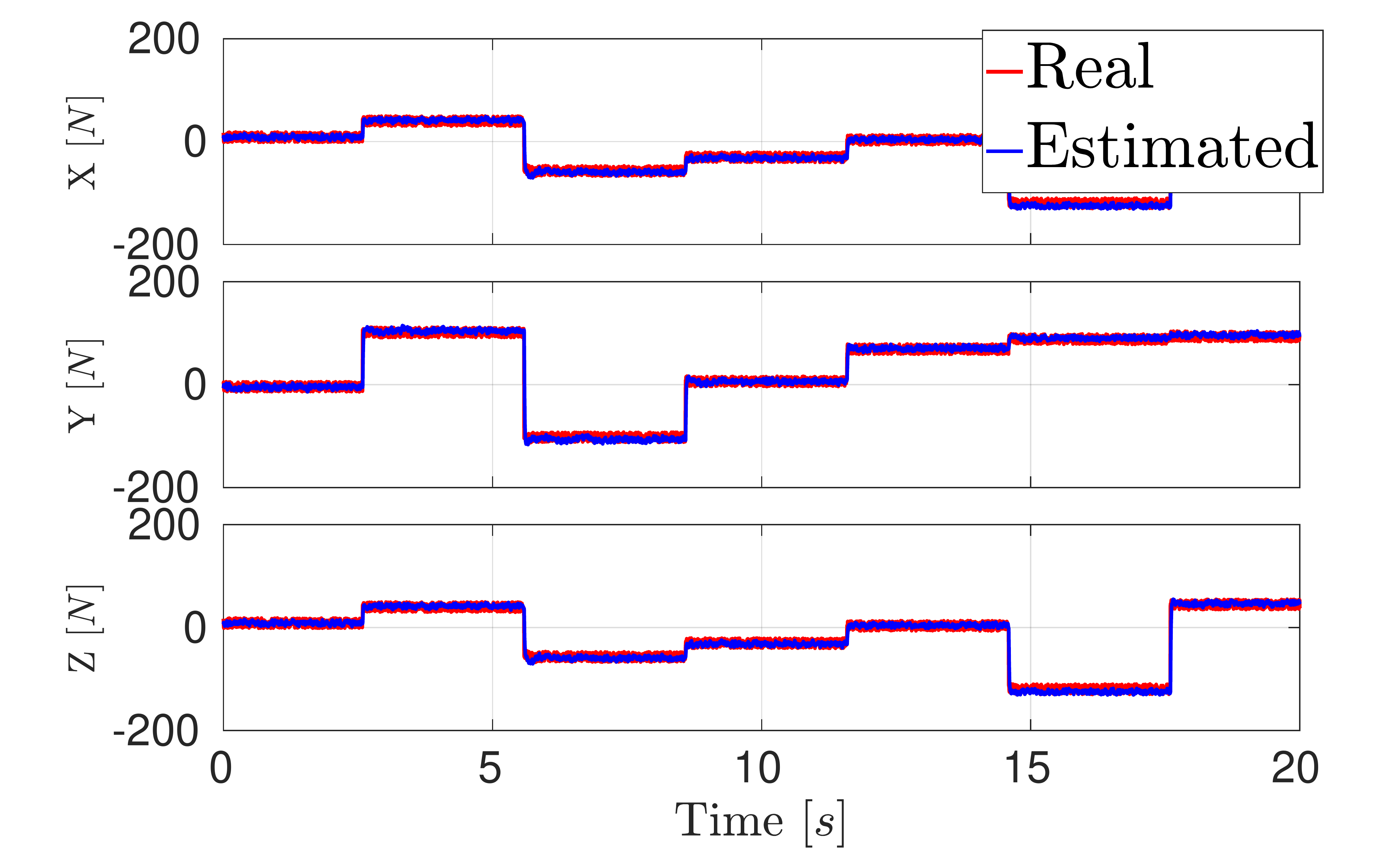}
	\caption{Simulated estimation of a time-varying external force. }
	\label{fig:forceEstimation}
\end{figure}

The result of the estimation is shown in Fig. \ref{fig:forceEstimation}:
the estimator is able to follow promptly the step changes with the gains set,
while a small filtering effect of the noise is  given by the observer
dynamics.

To evaluate the effectiveness of the \textit{compensation}, we
observe that an external force not properly compensated (\eg by
the Trunk Controller) results in \gls*{grf} which differ from the
desired ones (outputs
of the optimization), even in absence of modeling errors. This can create
slippages and loss of
contact, as shown in the accompanying
video\footnote{\gls*{mbdo} simulations:
\url{https://www.youtube.com/watch?v=_7ud4zIt-Gw&t=9m15s}}.

Therefore, a good metric to assess the effectiveness of the compensation is
the norm of the \gls*{grf} tracking error $\Vert f \Vert$.

\begin{figure}[bt]
	\centering

\includegraphics[width=0.8\columnwidth]{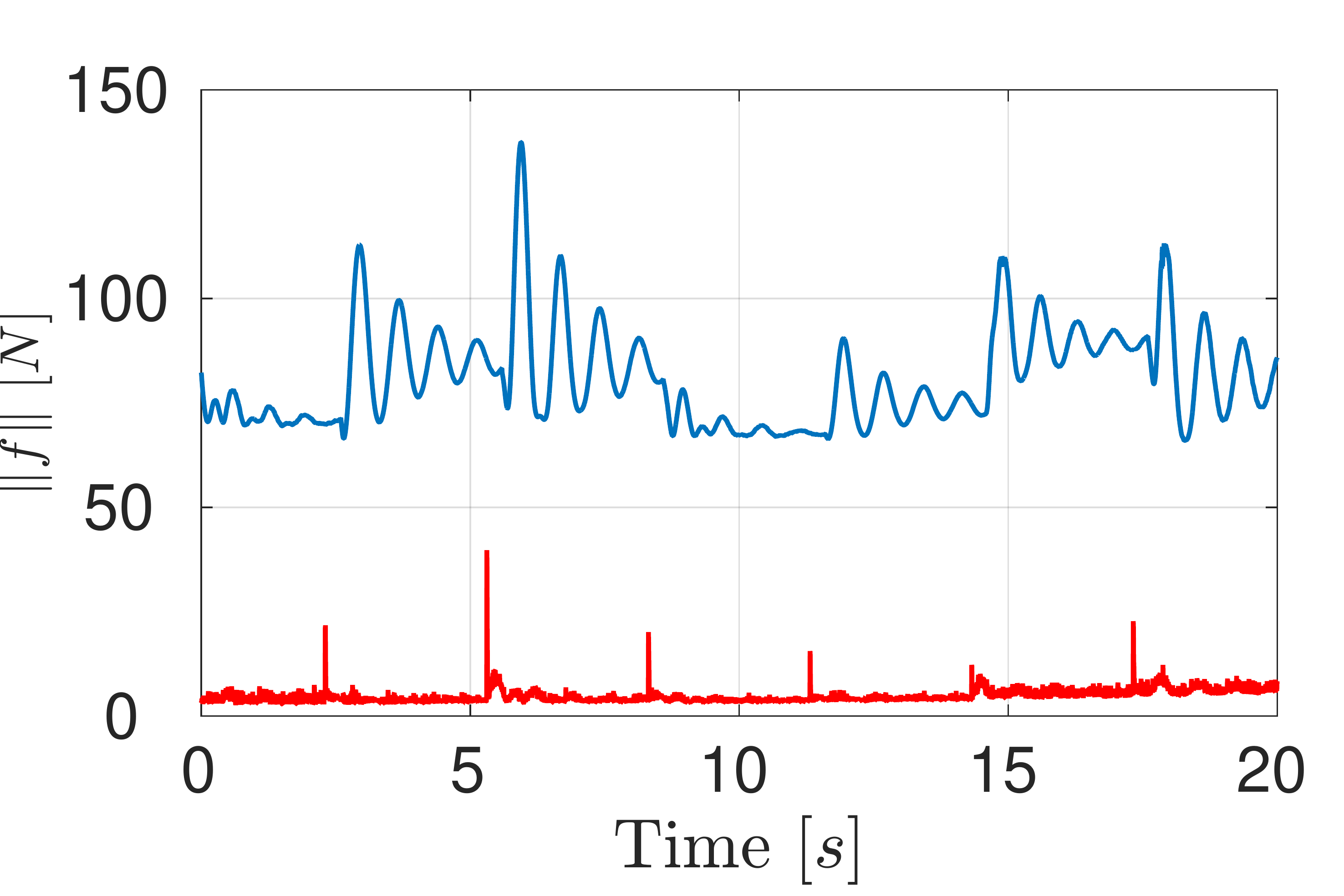}
	\caption{Norm of tracking error the with (red) and without (blue)
compensation of the external wrench. }
	\label{fig:forceCompensationSim}
\end{figure}

Figure \ref{fig:forceCompensationSim} shows that
the compensation improves significantly the \gls*{grf} tracking
by almost two orders of magnitude.

The video also shows a simulation of the robot pulling a wheelbarrow,
illustrating the \gls*{grf} (green arrows) and the location of the \gls*{zmp}
(purple sphere).
In the simulation the robot ``leans forward'', compensating for the
offset
in the \gls*{zmp}
created by the pulling force, while the back legs are more loaded (\ie
with larger
\gls*{grf}) with respect to the front ones,
due to the weight of the wheelbarrow.

\subsection{Experimental Results}
\label{sec:momentum-observer-experiments}
To demonstrate the effectiveness of our approach, we designed several test
scenarios in
which the robot is walking and compensating a time-varying disturbance
force\footnote{\gls*{mbdo} experiments:
\url{https://www.youtube.com/watch?v=_7ud4zIt-Gw&t=9m53s}}.
In all the experiments, we set the gains to $G_{\text{lin}} =
\diag(10,10,10),\;   G_{\text{ang}} =  \diag(1,1,1) $.

\subsubsection*{Experiment 1 - Pulling a Wheelbarrow}
pulling a wheelbarrow on a  ramp is an interestings experimental scenario for
our \gls{mbdo}. % because of the variety of the interaction force.
%Indeed, a wheelbarrow is sustained only on two
%wheels and to the robot. In this case,
%the external interaction with the robot has 
%have a vertical component that is mainly constant (gravity dependent).
%Moreover, there is a time-variant component due to
%the acceleration/deceleration of the inertia of the wheelbarrow itself. 
%
%As  additional difficulty,
%a robot walking on a \SI{22}{\degree} inclined ramp has to deal
%with a shrunk support polygon when performing stable locomotion.
%
\begin{figure}[tb]
\centering
\includegraphics[width=1.0\columnwidth]{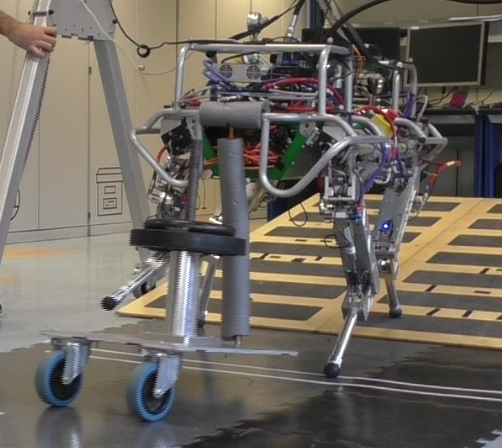}
\caption{HyQ quadruped robot pulling a wheelbarrow (of \SI{12}{\kilo\gram})
loaded
 with \SI{15}{\kilo\gram} of additional weight.
 The total vertical force acting on the robot is about \SI{13.5}{\kilo\gram}.
 The cart is attached to the robot through a rope.}
\label{fig:wheelBarrowWrenchExp}
\end{figure}

This task poses several challenges: 1) the wheelbarrow attached with a rope
(intermittent unilateral pull constraint)
creates  a
disturbance force which has a constant
vertical component (to counteract wheelbarrow gravity)
and a time-varying component (due to horizontal accelerations);
2) The motion of the wheel barrow results 
in potential unload of the rope that causes discontinuity in the force;
3) a walking gait involves a mutable contact condition, implying that the
compensation force must
be exerted by different legs at different times, without
discontinuitiess.
4) the additional loading of the wheelbarrow is done \textit{impulsively};
5) walking on a ramp shrinks the support polygon and
reduces the stability margin,
requiring a high accuracy in the estimation/compensation pipeline.

The accompanying video shows the robot walking on a ramp
while pulling a \SI{12}{\kilo\gram} wheelbarrow.
We performed different trials, with the disturbance wrench compensation enabled,
 adding \SI{10}{\kilo\gram} and \SI{5}{\kilo\gram} supplementary
weights (on top of the wheelbarrow), in different stages of the walk. 
Being the
total weight of \SI{27}{\kilo\gram} equally shared between the robot and the
wheels of the cart, we estimate the total load to be \SI{13.5}{\kilo\gram}.
With the compensation enabled,  the robot is able to smoothly climb the
ramp while  dragging this extra weight. Without the compensation, the robot was
struggling to
maintain the support polygon even with \SI{5}{\kilo\gram} of extra weight (for a
total load of approximately \SI{8.5}{\kilo\gram} hanging from the robot).
With \SI{15}{\kilo\gram} it was  unable to climb the ramp.

\begin{figure}[htb]
	\centering
	\includegraphics[width=1.0\columnwidth]{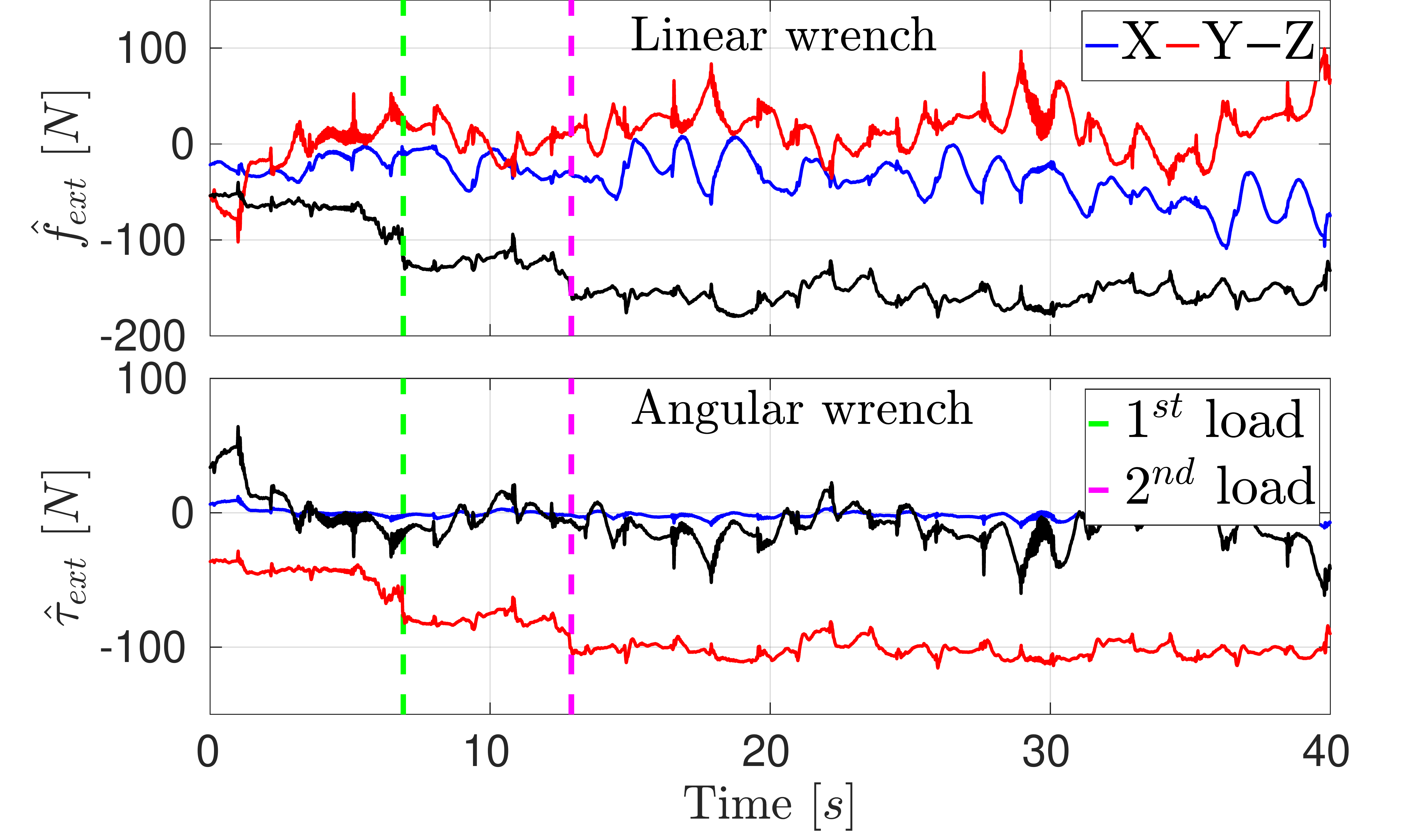}
	\caption{Wheelbarrow experiments. Estimated wrench in real experiments
while the
		robot carries out a wheelbarrow
		on rough terrain, (upper plot) linear part (lower plot)  the
		angular one. The two vertical lines represent 
		the moments where the 1-st (10 $kg$) and 2-nd (5 $kg$) load were
applied.}
	\label{fig:extWrenchBarrow}
\end{figure}

Figure \ref{fig:extWrenchBarrow} shows the components of the estimated wrench.
As expected, the most relevant ones are $f_{{ext}_x}$, which
is time varying, and $f_{{ext}_z}$, which is mostly the constant and vertical
component of the disturbance force. Note how the vertical component
$f_{{ext}_z}$ changes when two different loads  are added to the wheelbarrow.
The wheelbarrow is attached
to the back of the robot, hence it also exerts a constant negative moment
around the
$Y$ direction.

\begin{figure}[b]
\centering
\includegraphics[width=1.0\columnwidth]{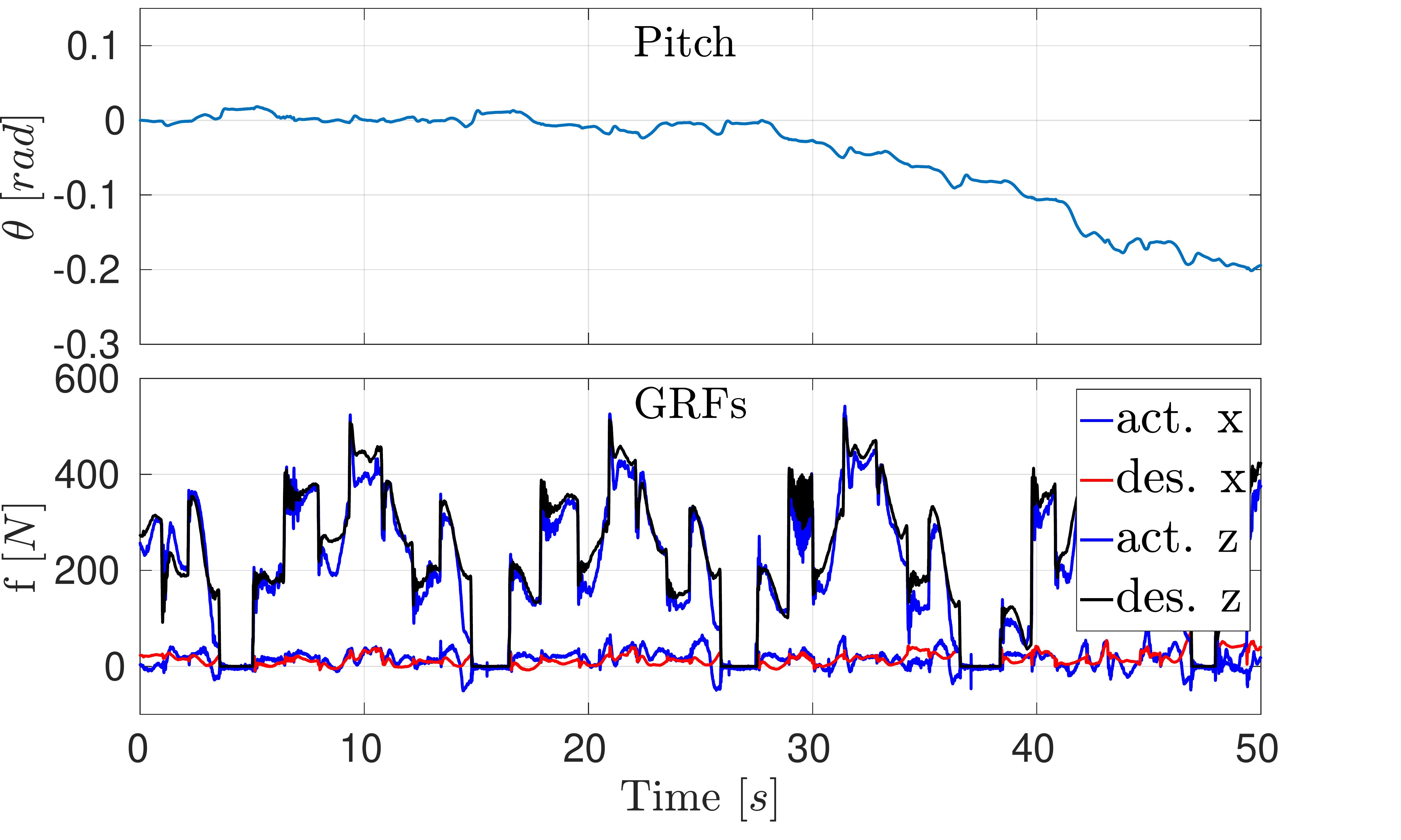}
\caption{Wheelbarrow experiments: (upper plot) trunk pitch angle and (lower
plot)
tracking of the $Z$ component of the \gls*{grf} of the $LH$ leg, 
while the robot carries the wheelbarrow
on ramp.}
\label{fig:grfBarrow}%\vspace{-1cm}
\end{figure}

Since the interaction force with the wheelbarrow
is unknown, again the  \gls*{grf} tracking is the only metric available to
assess the quality of the compensation.
Figure \ref{fig:grfBarrow} shows the pitch variation (upper plot)  while
walking first horizontally and then up the slope.
The \gls*{grf} tracking for the $LH$ leg is also
shown
(lower plot).
Thanks to the compensation, the tracking error is always below
\SI{40}{\newton} for the Z component (about \SI{5}{\percent} of the total
robot's weight) and \SI{20}{\newton} for the X component.
\subsubsection*{Experiment 2 -  Leaning Against a Pulling Force}
\begin{figure}
%\vspace{-4cm}
\centering
\includegraphics[width=1.0\columnwidth]{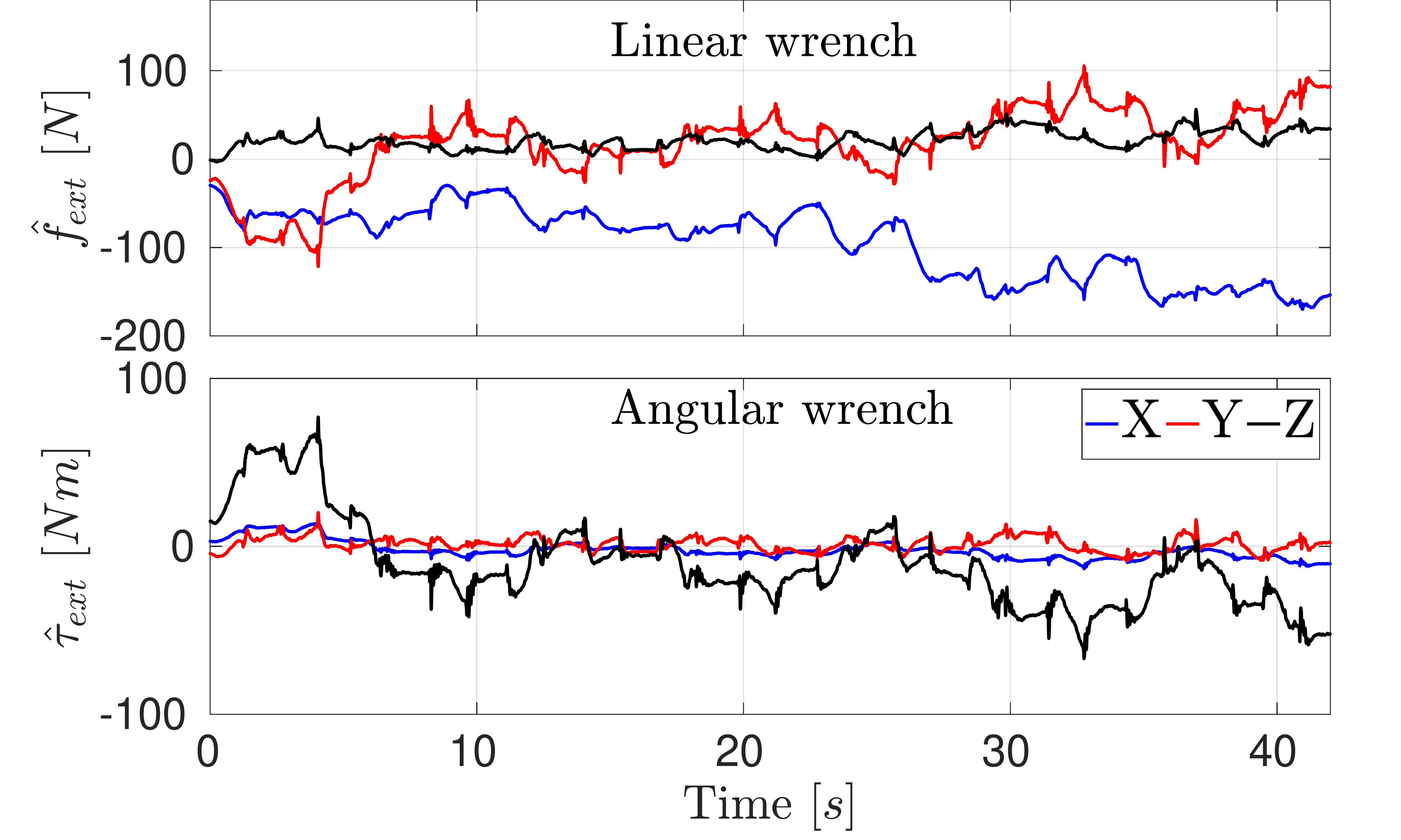}
\caption{Curling experiments. Estimated wrench: (upper plot) linear part, 
	the load is mainly along the $X$ component (horizontal) (lower
plot) angular part, all the moments are oscillating around zero, 
showing that the interaction force direction is passing close to the \gls{com}.}
\label{fig:curlingWrench}%\vspace{-3cm}
\end{figure}

In this experiment, a \SI{15}{\kilo\gram} load is attached to the back of
the robot with a rope. An
intermediate pulley transforms the
gravitational load into a horizontal force. This test allows to evaluate the
effectiveness
of the estimation/compensation pipeline in presence of
disturbance forces mostly \textit{horizontal}.
\begin{figure}
\centering
\includegraphics[width=1.0\columnwidth]{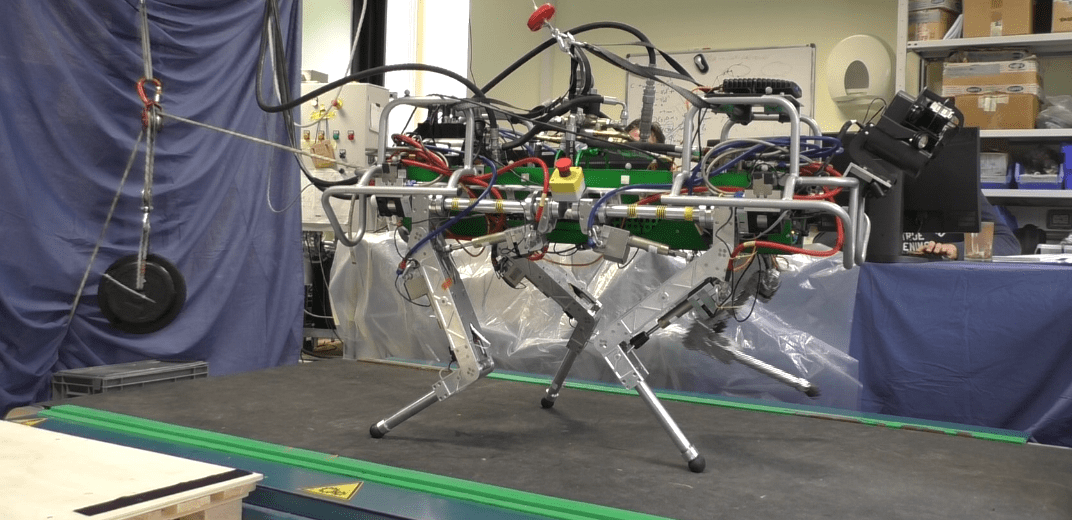}
\caption{HyQ quadruped robot dragging a $75N$ \textit{horizontal} disturbance
force (15 $kg$ load on a pulley) to testing the performances of the \gls{mbdo}.}
\label{fig:curlingWrenchExp}
\end{figure}
In the experiments, the robot is able to pull \SI{15}{\kilo\gram}
when walking with the compensation enabled. When the estimation is
disabled, the controller accumulates errors and suffers from instability.
Figure \ref{fig:curlingWrench} shows  that
the only significant component in the estimated wrench is along the $X$
direction (horizontal). The force increases, while the robot walks, because
the load is pulled up gradually (see video),
until it reaches the steady value of \SI{75}{N} (\ie the \SI{15}{\kilo\gram}
load halved by the pulley

).
The RViz visualization in the accompanying video shows
that the \gls*{grf} have a constant
component along the
$X$ direction (\eg pointing forward) to compensate for the backward pulling
force. At the same time,  when the compensation is active, the \gls*{zmp} is
shifted forward.
\section{Conclusions}
\label{sec:conclusions}
In this work we addressed the problem of locomotion on rough terrain.
We proposed a 1-step \textit{receeding horizon} heuristic planning strategy that can work with 
variable levels of exteroceptive feedbacks.

When no exteroceptive perception is available (and
no map of the environment is given to the robot)
the locomotion framework presented in this paper is still capable of
traversing challenging terrains.
This is possible thanks to its capability to \textit{blindly} adapt to the
terrain,
mainly because of the 1-step re-planning and of the \textit{haptic} touchdown.

When a map of the surrounding world is available, 
the proposed locomotion framework can exploit this information 
by including the actual terrain height and orientation in the planning, thus
decreasing the chances of stumbling and allowing to traverse more difficult
situations (\eg climbing stairs).

With the proposed approach we were able to achieve several
experimental results, such as traversing a template rough
terrain made of ramps, debris and stairs. 
We have also shown our quadruped  walking while
compensating for external \textit{time-varying} disturbance
forces caused by a load up to \SI{15}{\kilo\gram},
using the \gls*{mbdo} that we proposed. 
All these tasks where performed with the 
same  locomotion strategy that we presented in this paper.\\
Future works involve an extensive  analysis of the stability of 
the \gls{mbdo}, considering its interactions with the Trunk Controller loop.
Based on these results, we  intend to implement improved observer 
that maximizes the  stability region.
%- (locomotion on deformable mediums - sand, gravel, etc. - or abruptly - extensive experiments on the stair climbing
Our future research directions also include extending the navigation capabilities 
of HyQ to more complex environments made of rolling stones and big size
obstacles ($>$\SI{12}{\centi\meter} diameter). For these scenarios, we envision
the need to develop a shin collision detection algorithm capable of
kino-dynamic proprioception using a foot-switch sensor\cite{gao18patent}.
Finally, we plan to carry out extensive stair climbing experiments with the
proposed approach.
%
%If the step is too small there can be problems of shin collisions that must be overcame with
%reparative reflex or different strategies.
%In general we believe that the stepping strategy in the future should consider mobility issues.
%Also the stepping frequency can be changed in order to stay within the workspace (e.g. take smaller steps).
\section*{Appendix A}
\label{sec:appendix_A}
This appendix shows how to
compute the weighted average of $N$ orientation vectors.

Because finite rotations do not sum up as vectors do,
it is not possible to  apply ordinary laws of vector arithmetic to them.
%
%To do so first we recall the recursive expression for a linear (weighted) interpolation for a set of $N$ euclidean vectors $[v_1 v_N]$.
%For there vectors, averaging their Cartesian coordinate components is equivalent to drawing a line
%between them and picking a point on that line that is closer or further to each endpoint depending on that endpoint's weight.
%The  average $\bar{v}_{k+1}$ of the first $k+1$ vector can be written as a function of the previous average $\bar{v}_{k}$:
%\begin{align}
%\bar{v}_{k+1} = \bar{v}_{p} + t(v_{p+1} -  \bar{v}_{p}) \quad with \quad t \in [0 1]
%\label{eq:recursiveMean}
%\end{align}
%where $\bar{v}_k$ is the average of a subset of $k$ vectors  and \ref{eq:recursiveMean} is the parametric equation of the line from
%$\bar{v}_{p}$ to the new point $v_{k+1}$. Rearranging the equation we can express $\bar{v}_{k+1}$ as a weighted average of the  previous mean vector with the new vector:
%\begin{align}
%\bar{v}_{k+1} &= (1-t) \bar{v}_{k} + tv_{k+1}  = a_k\bar{v}_{k} + a_{k+1}v_{k+1}  \\
%%
%a_k &=\frac{ \sum\limits_{p=1}^k w_p}{ \sum\limits_{p=1}^{k+1}w_p} \quad
%%
%a_{k+1}  =\frac{w_{k+1}}{ \sum\limits_{p=1}^{k+1} w_p}\\
%%
%\label{eq:recursiveMean}
%\end{align}
%Note the weight $a_k$ of the previous average will be bigger  than the one $a_{k+1}$ of the new sample and $a_k+a_{k+1}=1$.
%
Specifically, the task of averaging orientations (described in the  algorithm \ref{geodesic}
written in pseudo-code) is equivalent to average points on a sphere,
where the line is replaced with a spherical geodesic
(arc of a circle)\footnote{The geodesic distance is the length of the shortest curve
lying on the sphere connecting the two points.}.

%Suppose $p(n)$ is the parametric function of that arc, with t ranging from $n_1$ to $n_2$ p(a) being the first vector and p(;) being the second vector. The weighted average $ p(a_1 n_1  + a_2)$.
%This is equivalent to apply a rotation
%towars $n_{2}$ starting from the orientation $n_1$. The angle of rotation will be scaled with the

\begin{algorithm}[ht]
	\caption{compute geodesic average}
	\label{geodesic}
	\begin{algorithmic}[1]
		\State  $w_1 \gets 1$
		\State  $\bar{n} \gets n_1$
		\For{$k=2$ to $N$}
		\State $\theta_k = acos(\text{dot}(\bar{n}, n_k))$
		\State $\bar{\theta}_k  =\theta_k \frac{ \sum\limits_{p=1}^{k-1} w_p}{ \sum\limits_{p=1}^{k}w_p}$
		\State $\bar{n} \gets$ rotate $n_k$ towards $ \bar{n}$ by angle $\bar{\theta}_k$
		\EndFor
	\end{algorithmic}
\end{algorithm}

The  iterative procedure illustrated above  is generic
and can be used to obtain the average of $N$ directional vectors $n_k$,
even though at each iteration we are only able to directly compute the average of two.
After the initialization, at each loop the actual average $\bar{n}$ is
updated with the next normal $n_k$.
To do so, we first compute the angle  $\theta_k$ between $n_k$
and the actual average $\bar{n}$. Then we scale this angle,
according to the (accumulated) weight of $\bar{n}$.
Finally,  $n_k$ is rotated  by the scaled  angle $\hat{\theta}$
toward $\bar{n}$ and the result is assigned back to $\bar{n}$.
%(see Fig.\ref{geodesic}).
%
%\begin{figure}[htb]
%\centering
%%\includegraphics[width=0.2\textwidth]{figures/geodesic.pdf}
%\caption{geodesic}
%\label{fig:geodesic}
%\end{figure}
\section*{Appendix B}\label{sec:appendix_B}
List of the main symbols used throughout the paper:\\
\begin{tabular}{@{}ll@{}}
	\textbf{Symbol} & \textbf{Description} \\
	$x_{com} \in \Rnum^3$ & coordinates of the CoM of the robot\\
	$x_{fi}, \dot{x}_{fi} \in \Rnum^3$ & positions and velocities of the $i^{th}$ foot\\
	$f_i \in \Rnum^{3}$ & contact forces of the $i^{th}$ stance foot\\
	$n$ & number of active joints of the robot\\
	$q^d_j, \dot{q}^d_j  \in \Rnum^n$ & joint reference positions and velocities\\
	$\tau_{ff}^d \in \Rnum^n$ & feedforward torque command\\
	$\tau_{pd}^d \in \Rnum^n$ & impedance torque command\\
	$\tau^d \in \Rnum^n$ & total reference torque command\\
	$\phi$ & roll angle of the robot's trunk\\
	$\theta$ & pitch angle of the robot's trunk\\
	$\psi$ & yaw angle of the robot's trunk\\
	$\Phi =  [\phi,
	\theta,\psi]$ & actual orient. of the robot's trunk\\
	$\Phi^d =  [\phi^{d},
	\theta^{d},\psi^{d}]$ & desired orient. of the robot's trunk\\
	$\Phi^d(0) = \Phi$ & des. orient. at the start of the move base\\
	$\Phi^{tg}$ & target orient. at the end of the move base\\
	$x_{com}^{tg}$ & target CoM position of the trunk\\
	$x_{{com}_p}^{tg}$ & projection of $x_{com}^{tg}$ on the terrain plane\\
	$h_r$ & robot's height\\
	$f_{ext}, \tau_{ext}$ & external disturbance force and torque
\end{tabular}

\section*{Acknowledgement}
\thanks{This work was supported by Istituto Italiano di Tecnologia (IIT),
	with additional funding from the European Union's Seventh Framework Programme for research,
	technological development and demonstration under grant agreement no. 601116 as part of
	the ECHORD++ (The European Coordination Hub for Open Robotics Development) project
	under the experiment called \textit{HyQ-REAL}.}

\small
\bibliographystyle{ieeetr}
\bibliography{references/references}
\end{document}